\documentclass[journal,onecolumn]{IEEEtran}

\usepackage[colorlinks,urlcolor=blue,linkcolor=blue,citecolor=blue]{hyperref}

\usepackage{color,array}
\usepackage{wrapfig}
\usepackage{graphicx}

\usepackage{amsmath}
\usepackage{amsfonts}
\usepackage{amssymb}
\usepackage{mathrsfs}
\usepackage{xcolor}
\usepackage{comment}

\usepackage[table]{xcolor}
\usepackage{array, booktabs, multirow}
\usepackage{enumitem}
\usepackage{caption}

\newcolumntype{L}[1]{>{\raggedright\arraybackslash}m{#1}}

\usepackage{blindtext}

\usepackage{MnSymbol}

\usepackage[cal=boondox,scr=boondoxo]{mathalfa}

\usepackage{float}
\usepackage{subfig}
\usepackage{fancybox,graphicx}
\usepackage{subfig}
\usepackage{caption}
\usepackage{color}
\usepackage{authblk}

\usepackage[colorlinks]{hyperref}

\usepackage{accents}
\usepackage{cite}

\begin{document}

\title{Artificial Human Intelligence:
The role of Humans in the Development of Next Generation AI} 

%\author{\textit{Authors are anonymized for review.}}

%\author{Suayb S. Arslan}

\author{Suayb S. Arslan%
\thanks{
Author is affiliated with the Department of Computer Engineering, Boğaziçi University, Istanbul, Türkiye, and the Department of Brain and Cognitive Sciences, Massachusetts Institute of Technology (MIT), Cambridge, MA, USA. 
Emails: \href{mailto:suayb.arslan@bogazici.edu.tr}{suayb.arslan@bogazici.edu.tr}, 
\href{mailto:sarslan@mit.edu}{sarslan@mit.edu}.

This work has been accepted for publication in the IEEE Transactions on Emerging Topics in Computational Intelligence. Please contact IEEE for the most up-to-date version following further editorial revisions.

© 2025 IEEE. Personal use of this material is permitted. Permission from IEEE must be obtained for all other uses, in any current or future media, including reprinting/republishing this material for advertising or promotional purposes, creating new collective works, for resale or redistribution to servers or lists, or reuse of any copyrighted component of this work in other works.
}%
\thanks{}\\
}

%\footnote{S. S. Arslan is with the department of computer engineering at Bogaziçi University, Bebek, Istanbul, 32324 Turkey and serves as a Professor and holds an affiliate position at the department of brain and cognitive sciences at MIT, Cambridge, MA 02138 USA (e-mail: suayb.arslan@bogazici.edu.tr, sarslan@mit.edu). He is also directing The institute for Data Science and Artificial Intelligence (DSAI) at Bogaziçi University, Bebek, Istanbul, Turkey.}

%\thanks{S. S. Arslan is with the department of computer engineering at Bogaziçi University and holds an affiliate position at the department of brain and cognitive sciences at MIT, Cambridge, MA 02138 USA (e-mails: suayb.arslan@bogazici.edu.tr, sarslan@mit.edu).}
%}

%\thanks{S. B. Author, Jr., was with Rice University, Houston, TX 77005 USA. He is now with the Department of Physics, Colorado State University, Fort Collins, CO 80523 USA (e-mail: author@lamar.colostate.edu).}
%\thanks{T. C. Author is with the Electrical Engineering Department, University of Colorado, Boulder, CO 80309 USA, on leave from the National Research Institute for Metals, Tsukuba, Japan (e-mail: author@nrim.go.jp).}
%\thanks{This paragraph will include the Associate Editor who handled your paper.}

%\markboth{Journal of IEEE Transactions on NNLS}
%{Suayb S. Arslan \MakeLowercase{\textit{et al.}}: Bare Demo of IEEEtai.cls for IEEE Journals of IEEE Transactions}

\maketitle

\begin{abstract}
Human intelligence, the most evident and accessible form of source of reasoning, hosted by biological hardware, has evolved and been refined over thousands of years, positioning itself today to create new artificial forms and preparing to self--design their evolutionary path forward. Beginning with the advent of foundation models, the rate at which human and artificial intelligence interact with each other has exceeded any anticipated quantitative figures. The close engagement led both bits of intelligence to be impacted in various ways, which naturally resulted in complex confluences that warrant close scrutiny. Recent advances, such as DeepSeek, exemplify this interplay: the novel contributions, we argue, draw indirect inspiration from biological principles like modular neural specialization and sparse episodic encoding, addressing computational bottlenecks while aligning with human-inspired scalability. In the sequel,  using a novel taxonomy, we shall explore this interplay between human and machine intelligence, focusing on the crucial role humans play in developing ethical, responsible, and robust intelligent systems. We briefly delve into various aspects of implementation inspired by the mechanisms underlying neuroscience and human cognition. In addition, we propose future perspectives, capitalizing on the advantages of symbiotic designs to suggest a human-centered direction for next-generation  developments, focusing on the augmentation role of AI. We finalize this evolving document with some thoughts and open questions yet to be addressed by the broader community. 
\end{abstract}

%\begin{IEEEImpStatement}
%Intelligence is a notion highly debated on its definition and implementation. It is quite central to inquire the role of humans in the development of AI and distinguish the similarities and differences in the state-of-the-art techniques. This work explores the intricate and evolving relationship between human and artificial intelligence, emphasizing the critical role humans play in guiding the development of ethical, responsible, and robust intelligent systems by introducing novel categorizations and a taxonomy. Drawing on insights gained through years and extensive exploration of current research, this study highlights a specific human-centered approach to next-generation AI, inspired by the mechanisms of neuroscience and human cognition. By examining the symbiosis between human and machine intelligence, it emphasizes the need for continuous reflection and adaptation as we shape the future of AI, encouraging the broader community to engage with unresolved questions and challenges in this dynamic and evolving field.
%\end{IEEEImpStatement}


%\tableofcontents

\section{Introduction} \label{section1}

\IEEEPARstart{I}{n} the wake of the Artificial Intelligence (AI) revolution, industries and societies, more broadly almost everyday life errands have undergone profound transformations. The relentless evolution of AI technologies and digital learning mechanisms prompts critical inquiries into the extent of human contribution in steering these advancements. While modern AI systems showcase unparalleled prowess in automation, decision-making, and closed-domain problem-solving, the indispensability of human engagement and intervention remains unequivocal in shaping the trajectory of next-generation AI systems as it has been the source of inspiration since the early appearances in the past \cite{russell2016artificial}.

\begin{figure}
\vspace{-4mm}
  \begin{center}
  \includegraphics[width=0.7\textwidth]{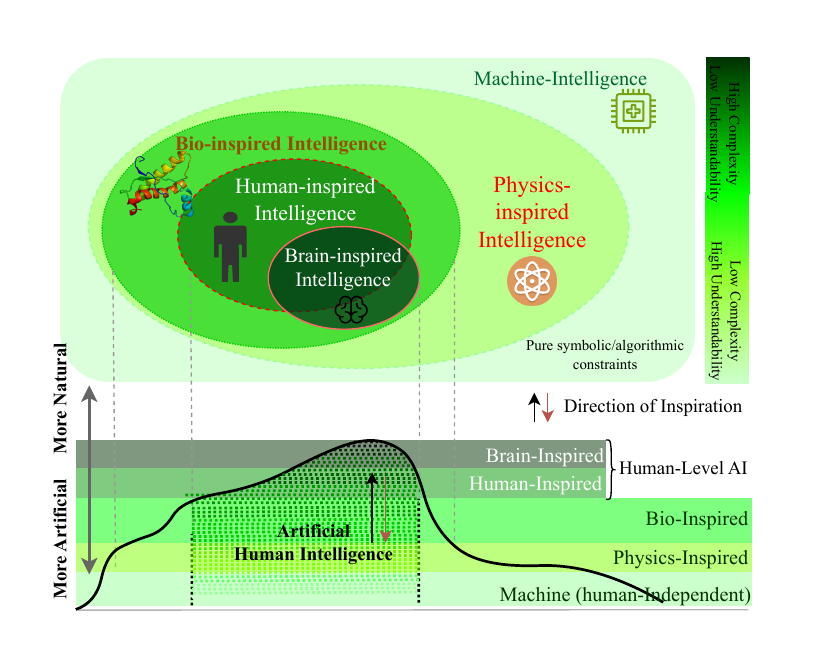}
  \end{center}
  \caption{
\textcolor{gray}{The kinds of inspiration sources in search of modeling intelligence with their corresponding levels of complexity and explainability within the context of \textit{Artificial Human Intelligence}, a concept that bridges (through inspiration) between brain-inspired and human-independent intelligences.}}
    \label{fig:IntelTypes}
\end{figure}

The complexity of the intelligence problem, to date still resisting a universally accepted definition, often motivates adaptation of the human model in identifying intelligent behavior. AI, in most forms conceivable, traces its origin with either mere mimicking of human functionality or a direct product of human intellect, inspired from various sources such as the outside physical world. This dependence inherently biases its design. The taxonomy briefly illustrated in Fig. \ref{fig:IntelTypes} reveals various sources of inspiration for human-level AI development, ranging from bio--inspired and brain-inspired {(mostly implemented with connectionist structures)} to purely symbolic algorithmic constraints. The same illustration clearly marks the novel concept of \textit{Artificial Human Intelligence} (AHI) that spans all layers of inspiration and impacts the design evolution of next-generation AI around human-centric skills. Human-inspired intelligence, as a subclass of AHI, acts as a critical category that will actually interlink human cognition with a wide variety of machine capabilities. Fostering collaboration among these domains could enable humankind to harness revolutionary cognitive technologies and inventions that push the boundaries {further}. Moving through multiple layers of inspiration, it forms a synergistic feedback loop that will yield human-level AI, representing the emerging relationship between digital and biological computations.

In this brief exposition, our objective is to investigate the strategic placement of human--centered inquiries in the advancement of contemporary AI systems despite the recent remarkable (not necessarily human--centric) developments in Machine Learning (ML) and Computer Vision (CV) fields \cite{jones2024ai}. 
Considering the strategies of machine and human intelligences in completing various tasks and their complementarity in behavior, a potential fusion of both emerges as a plausible research direction that might soon become an applicable reality.  Due to the involvement of humans as biological organisms, more specifically brain-centric investigations, to be implemented on pure algorithm-driven computational platforms, we have decided to label this specific fusion of intelligences and all the interactions within as AHI and explicate various intriguing topics of discourse for the mutually beneficial association of the two (seemingly) mechanistically distinct notions \cite{sgantzos2024minds}.

%Naturally, the exploration of both forms of intelligence necessitates addressing numerous intriguing questions. These include early human brain developments and their connections to neuroscience, as well as biological hardware dependencies, efficiency of implementations and the potential long-term effects on both human and machine intelligences as both intelligent forms will likely coexist in any foreseeable future. 

\begin{figure*}[t!]
    \centering
    \includegraphics[width=6.5in]{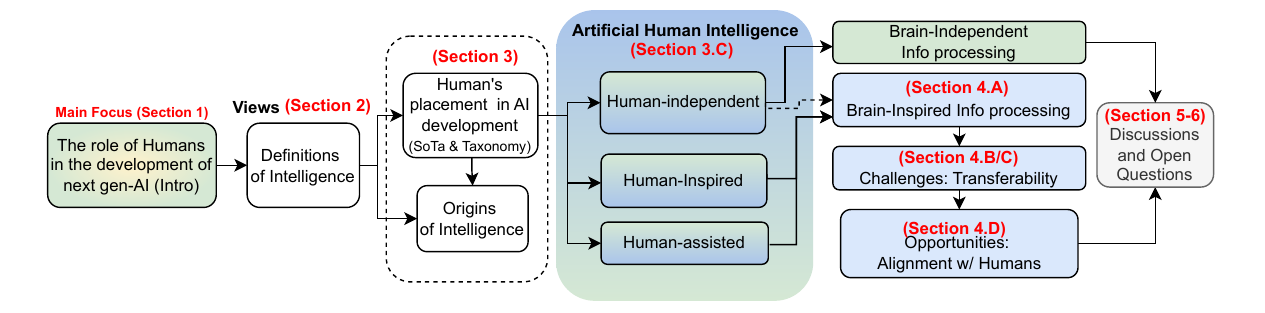}
    \caption{
\textcolor{gray}{Summary of the manuscript detailing various topics of interest and where they are covered with associated section and/or subsection numbers.}}
    \label{fig:Summary}
\end{figure*}

{The organization} of the paper is {summarized} in Fig. \ref{fig:Summary}. {The main motivation} is introduced in Section \ref{section1}. Section \ref{section2} provides some of the expert opinions on the subject. In Section \ref{section3}, the origins of biological intelligence {is discussed}, later followed by the introduction of the novel notion of AHI in Section \ref{section3.2}, arguing in favor of the confluence of artificial and natural intelligences. {We hypothesize that this unique interaction may become the standard for future human–human and human–machine communication. This should not be viewed as an alternative to Artificial General Intelligence (AGI) \cite{goertzel2014artificial,goertzel2007artificial,bubeck2023sparks}, but rather as a step towards developing more general and capable machine intelligence with human-like behavioral, architectural, and functional features \cite{korteling2021human}. Whether or not this effort would lead to AGI is an open question. Section \ref{section5} examines human-level AI, focusing on the role of humans in the emerging AI ecosystem, related challenges, and alignment issues. Section \ref{section6} considers the potential effects of AI on the human brain and cognition, along with the long-term management actions, before concluding in Section \ref{concSection}.}  %This unique interaction, we hypothesize, would be the de facto standard of future human-to-human as well as human-to-machine communications and interfaces. It must be emphasized though that such an interaction is not an alternative to Artificial General Intelligence (AGI) \cite{goertzel2014artificial,goertzel2007artificial, bubeck2023sparks} in any sense, but rather a tentative direction {towards} achieving more generic and capable machine intelligence with human-like features at behavioral, architectural and functional levels \cite{korteling2021human}. 
%Later, we provide details about what we mean by artificial human intelligence and review few attempts as the state-of-the-art (SoTA) in this field (Section \ref{section4}). We further use brain-centric categorization of past efforts and consider the missing elements in brain-inspired information processing. 
%In Section  \ref{section5}, we delve into human-level AI, some of the questions that concern where exactly humans will be positioned in the newly formed AI ecosystem, addressing few challenges and discuss the final alignment with humans. Finally in Section \ref{section6}, we comment/speculate on the effects of AI on the human brain and cognition and the set of required management actions in the long term as part of our discussions, before we conclude the paper in Section \ref{concSection}. %We finally conclude with some of the most compelling open questions relevant to our subject matter. 

\section{Views on Intelligence} \label{section2}

{Although often treated as straightforward in everyday use, the concept of intelligence is highly complex. Rapid learning, lifelong adaptation, knowledge representation, abstract reasoning, symbolic problem-solving, and creativity all contribute to its elusive nature. Psychologists have long debated its definition and measurement, fueling controversies over human valuation and performance assessment, particularly in relation to the current education system \cite{salomon1991partners}.}
%The notion of intelligence, although seemingly straightforward in everyday language to define or use, is remarkably sophisticated and complex. Attributes like expeditious learning or life-long adaptation, knowledge accumulation and representation, abstract reasoning, symbolic problem-solving, and creativity contribute to its elusive and unfathomable nature. Psychologists have long debated its definition and measurement, sparking controversies regarding human valuation and performance assessment, particularly relevant and applicable to {the} current education system \cite{salomon1991partners}. 
Despite some consensus on measuring intelligence via standardized tests \cite{terman1916uses}, issues persist around their scope, potential biases, and the spectrum of skills they believe these experiments assess. Concurrently, the discourse on machine intelligence faces even greater complexity due to the diverse forms and abilities of machines. Defining intelligence in precise terms, applicable across a wide range of systems, poses significant challenges, especially as tasks once considered benchmarks of human intellect become now remarkably trivial for machines (like ImageNet \cite{deng2009imagenet} or EcoSet \cite{mehrer2021ecologically} based categorization tasks). As technology evolves, so too does our perception of intelligence, necessitating a comprehensive and enduring definition that transcends sensory limitations, environmental constraints, and specific hardware types, yet remains practically achievable.

%The quest for defining intelligence is a multifaceted challenge, labyrinthine in its nature and diverse in its interpretations. 
Intelligence includes a spectrum of cognitive abilities, adaptive behaviors, and problem-solving skills exhibited by living organisms. At its core, intelligence involves the capacity to acquire knowledge, reason abstractly ({e.g.} counterfactual thinking), solve problems, learn from exploratory experience ({in 3D} world model), and adapt to unseen circumstances (generalization) quickly without necessarily using many examples ({e.g.} one-shot learning \cite{vinyals2016matching}). {Intelligence manifests in diverse forms across species, each adapted to its ecological niche and evolutionary context. Human intelligence, often seen as the pinnacle of cognition, is marked by complexity, versatility, and symbolic thought, language, and culture.}
%Intelligence might manifest itself in diverse forms across species, each tailored to suit the ecological niche and evolutionary context of the organism. Human intelligence, often hailed as the pinnacle of cognitive evolution, is characterized by its complexity, versatility, and capacity for symbolic thought, language, and cultural and societal intercourse. 
However, defining human intelligence as the ultimate form of intelligence risks anthropocentric bias and overlooks the diverse manifestations of intelligence found in other organisms \cite{anderson2010octopus}. %vallor2017artificial

{Russell \cite{russell2022human} defines intelligence as an entity’s ability to act in ways it expects will achieve its goals based on its perceptions. Applied to machines, AI thus becomes the design of systems that optimize given objectives. However, Russell argues that this one-dimensional, “unary” view is insufficient for capturing the complexity of human intelligence.} %Russell \cite{russell2022human} suggests that intelligence can be manifest as an entity’s ability to take actions that it expects will help achieve its goals provided what it has perceived. In his view, when we apply this definition to machines, AI essentially becomes about building systems that optimize for specific objectives given as inputs.  Russell’s framework suggests that the common, one-dimensional view of intelligence is incomplete when we consider human intelligence. This “unary” definition might work for an autonomous agent pursuing its own goals and taking actions, but human intelligence is far more complex. 
%{Scholars have proposed models to explain this multidimensional nature of intelligence. Gardner's theory of multiple intelligences \cite{davis2011theory} broadens the concept to distinct modalities—linguistic, logical-mathematical, spatial, musical, bodily-kinesthetic, inter-personal, intra-personal, and naturalistic—each reflecting unique talents and capabilities. More recently, deep learning has enabled constrained yet diverse forms of intelligence that perform complex tasks and occasionally display human-like cognition \cite{liao2016bridging}. This pluralistic view invites rethinking developmental trajectories with cross-modal interactions. Still, replicating the full scope of human intelligence remains elusive, as current systems excel only in narrow domains and falter at generalization, \textit{common sense}, and contextual understanding. Moreover, ethical issues—autonomy, accountability, and bias—underscore the societal stakes of emerging machine intelligences, demanding careful ethical, social, and philosophical reflection.}
{Scholars have proposed models to capture the multidimensional nature of intelligence. Gardner’s theory of multiple intelligences \cite{davis2011theory} extends the concept to distinct modalities—linguistic, logical--mathematical, spatial, musical, bodily--kinesthetic, interpersonal, intrapersonal, and naturalistic--each reflecting unique capabilities. More recently, deep learning has enabled constrained yet varied forms of intelligence that perform complex tasks and sometimes exhibit human-like cognition \cite{liao2016bridging}. This pluralistic view invites rethinking developmental trajectories with cross-modal interactions. Yet, replicating the breadth of human intelligence remains elusive, as current systems excel only in narrow domains and falter at generalization, common sense, and contextual reasoning. Ethical concerns--autonomy, accountability, and bias--further underscore the societal stakes, requiring careful reflection.}

%Scholars have proposed various models and frameworks to elucidate the multidimensional nature of intelligence. Gardner's theory of multiple intelligences \cite{davis2011theory} posits that the broad notion of intelligence should comprise distinct modalities, including linguistic, logical-mathematical, spatial, musical, bodily-kinesthetic, inter-personal, intra-personal, and naturalistic intelligences. Each intelligence represents a domain of expertise and proficiency, offering insights into the diverse talents and capabilities of individuals.  On the other hand, recent developments in deep learning have enabled  \textit{rather constrained but various forms of intelligences} capable of performing complex tasks and sometimes exhibiting human-like cognitive abilities \cite{liao2016bridging}. This pluralistic account of intelligence allows a rethinking of developmental trajectory with necessary cross modality interactions. However, replicating the full breadth and depth of human intelligence remains elusive, as deep learning systems often excel in narrow domains but struggle with generalization, common-sense reasoning, and context-dependent understanding. Furthermore, ethical considerations surrounding the learning systems raise concerns about the societal impact of new forms of intelligence, including issues of autonomy, accountability, and bias. As we venture into the realms of various forms of machine intelligence, navigating these challenges requires careful consideration of ethical, social, and philosophical implications.

Brooks highlights that in the early stages of AI research, intelligence was defined as tasks that highly educated scientists found challenging, such as \textit{chess}, proving mathematical theorems or solving complex problems \cite{brooks2018intelligence}. %\cite{brooks2003flesh} 
Tasks that young children effortlessly accomplished, like visually distinguishing between objects (e.g., a coffee cup) {or} walking on two legs without losing balance were not initially regarded as activities requiring intelligence. {It is argued} that intelligence should be regarded as the everyday activities humans constantly perform. Winston \cite{winston1984artificial} delineates the 
objectives of AI as twofold: The 
development of practical intelligent systems and the 
comprehension of human intelligence, arguing in favor of a mutually beneficial enterprise. {Legg and Hutter \cite{legg2007collection} suggest that, despite {the diversity}, a formal encompassing universal intelligence definition may be possible, closely linked to the theory of optimal learning agents \cite{hutter2005universal}. Such agents are designed to learn and act to maximize performance in achieving goals, employing algorithms that enhance learning and decision-making through interaction and feedback.}

\section{Path leading to AHI}
\label{section3}

\subsection{Early Origins {and AHI Definition}}

{Neither the origin of biological intelligence nor a theory for it has been adequately addressed in the past. In fact, all definitions of intelligence concerning origins contain a strong subjective component. Whether Turing or Newell \cite{newell1982knowledge}, their descriptions—knowledge-level considerations, beliefs, and intentions to a system—are grounded in human judgments, making them subjective. These approaches also frame emergence as discrete steps toward complex intelligent behavior rather than a gradual course. By contrast, Steels \cite{steels1996origins} proposes a continuum: a construct evolving from chemical systems to living organisms. Close inspection of complex adaptive systems reduces the quest for origins to two questions: \textbf{(1)} how the brain demonstrates behavioral plasticity and \textbf{(2)} how a symbolic layer have emerged early in development \cite{steels1996origins}.} %weng2011symbolic

%Neither the origin of biological intelligence nor a theory for it has been adequately addressed {in the past}. In fact, we primarily notice that all definitions of intelligence leading to origins seem to have a strong subjective component.  Whether it be Turing or Newell \cite{newell1982knowledge}, their descriptions of intelligence, involving knowledge-level considerations, beliefs, and intentions to a system, are strictly based on human judgments, making both of them subjective. In addition, these approaches adapt discrete steps of emergence towards complex intelligent behavior, rather than a gradual time course. On the other hand, Steel \cite{steels1996origins} supports an alternative approach which highlights the continuum nature: A construct progressively evolving from chemical systems to actual living organisms. Through close inspection of evolving complex adaptive systems, the quest for biological origins can be reduced down to addressing two fundamental questions, namely,  \textbf{(1)} how the brain demonstrates its remarkable behavioral plasticity and \textbf{(2)} how a symbolic layer might have emerged early in the development \cite{steels1996origins, weng2011symbolic}. 

\begin{figure*}[t!]
    \centering
    \includegraphics[width=7.2in]{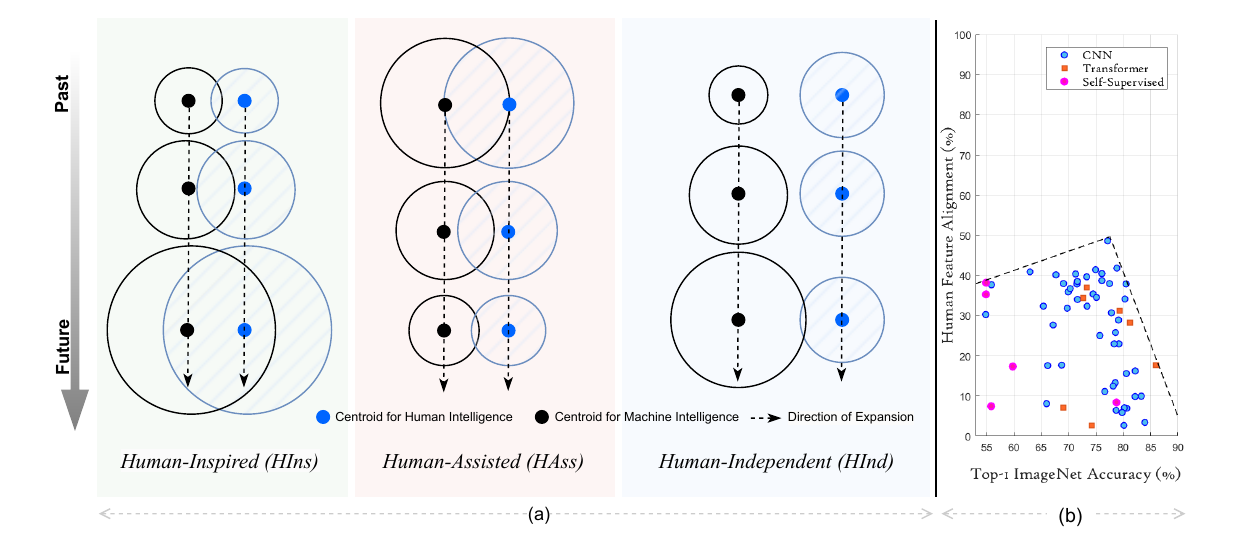}
    \caption{
\textcolor{gray}{ \textbf{(a)} In \textit{HIns} Intelligence, developments in machine intelligence are primarily driven by insights from biology and neuroscience, leading to overlapping advancements in both fields. Conversely, in \textit{HAss} intelligence, the overlap tends to diminish over time, resulting in more specialized assistance from either humans or machines, fostering the development of hybrid systems. Finally, in \textit{HInd} intelligence, the intrinsic features of machine intelligence evolve freely, naturally causing a divergence in the fundamental principles governing human and machine intelligence. (b) A measurement of human feature alignment with respect to machines (CNNs, Transformers, Self-Supervised) as a function of accuracy \cite{fel2022harmonizing}.}}
    \label{fig:intelX}
\end{figure*}

Biological intelligence most likely emerged from active perception and tight/attentive mechanisms that entail 3D interactions with real-world phenomena through multi-sensory input \cite{pfeifer2006body, marcus2020next}. In other words, biological intelligence requires to exist in multi-dimensional feedback-enabled open environments to emerge and develop \cite{marcus2020next}, which eventually manifests agency-like behaviour. %The design of intelligent machines can potentially provide clues about origins as they might likely simulate the emergence. 
{When developing such models, one may consider various levels of natural intelligence}\footnote{According to Gardner, naturalistic intelligence is the ability to identify, categorize, and engage with environmental elements like objects, animals, and plants. This enables differentiation between species, groups, or objects and understanding their ontological connections.} \cite{davis2011theory}. {Human intelligence, being immediately observable, has long informed machine-intelligence research—particularly the deep learning community—driving the design of novel architectures, training methodologies, and learning techniques. Multimodal sensory processing is fundamental to human intelligence, with considerable evidence for its joint development in early childhood \cite{gori2012visual}}. %petrini2014vision
%but one of the observable models immediately accessible and intriguing is the human intelligence which has been heavily leveraged by machine intelligence innovators (particularly deep learning community) in the past for so many years to inspire novel architectures, training methodologies, learning techniques, etc. Multi-model sensory input processing is essential to human intelligence as there is considerable evidence to support their joint progression early in childhood \cite{petrini2014vision,gori2012visual}. 
In fact, recent work on the computational modelings of visually guided haptic sensory systems advocates for the value of multimodal fusion to improve the visual experience and performance \cite{wan2020artificial}. As different sensory systems can be linked to distinct intelligent behaviors, a growing number of studies suggest that their synchronization and fusion lead to improved accuracy in the completion of specific tasks \cite{gao2016deep}, which demonstrates the potential for integrating more general and non-biological intelligences. 
One of the most compelling pieces of evidence supporting this finding is the success of multimodal FMs, which showed that a missing modality can be compensated by the presence of other modalities \cite{wu2023pi}. 

We characterize AHI to describe trichotomous, consisting of three kinds of machine intelligences,
{
\begin{itemize}
    \item \textit{\textbf{Human-Inspired (HIns):}} Designs whose core architectures, inductive biases, or learning rules are derived from human cognition or neurobiology (e.g., spiking dynamics, local learning, developmental curricula). \textit{Key criterion:} inspired-by human mechanisms — does not require humans at runtime. %AI systems that draw inspiration from human cognitive processes, such as learning, reasoning, or problem-solving, to mimic human-like decision-making and understanding.
    \item \textit{\textbf{Human-Assisted (HAss):}} Systems that operationally depend on humans — their training, tuning, or real-time decisions require explicit human input, feedback, or supervision (e.g., RLHF). \textit{Key criterion:} human involvement is integral to performance or safety. %AI systems that rely on human input, collaboration, or feedback for operation, training, or improvement, often combining human expertise with machine efficiency.
    \item \textit{\textbf{Human-Independent (HInd):}} Systems engineered to learn, adapt, and act autonomously with minimal human intervention, using self-supervision, automated data collection, or model search. \textit{Key criterion:} autonomy in learning/operation; humans only set goals or monitor. %capable of operating, learning, and making decisions without direct human involvement, designed to perform tasks independently.
\end{itemize}
}

%Just like found in the integration of multimodal sensory inputs, this paradigm's objective is to improve system performance by leveraging the distinct learning mechanisms of both humans and various AI approaches, which might result in complementarity in processing and task performance.

%Since each one of these trejectories are rooted based on human capabilities and human intelligence, we coined the name AHI to describe all. 
 %Additionally, the involvement of humans or understanding tasks performed by the human brain can help regulate ethical considerations, symbolic computations, and emotional aspects of the underlying functional systems. 

\begin{figure*}[t!]
    \centering
    \includegraphics[width=5.5in]{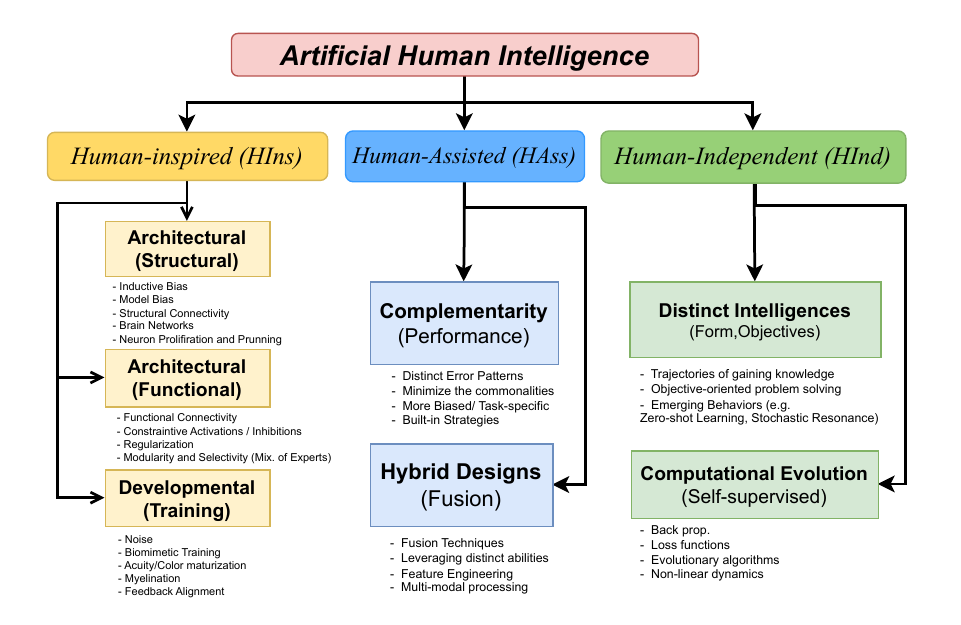}
    \caption{\textcolor{lightgray}{Categorization of Human-centered intelligences and associated subcatagories.}}
    \label{fig:AHIplot}
\end{figure*}

\subsection{The current SoTA and Taxonomy}
\label{section3.2}

{Current AI systems place limited emphasis on defining the origins of intelligence and often diverge from the objectives of human or machine intelligences. Early work pragmatically modeled human intelligence through layered mind models \cite{rescorla2015computational}, later reflected in Convolutional Neural Networks (CNNs) \cite{krizhevsky2017imagenet}, where hierarchical cortical structures inspired inductive biases. Intelligence was typically measured by efficiency and accuracy relative to humans. However, due to the field’s heavy emphasis on scaling (bigger models + more data), lower built-in inductive biases, etc., machine intelligence evolves along trajectories not necessarily compatible with human development. Studies such as \cite{fel2022harmonizing} show that as accuracy improves, alignment with human features declines. Popular architectures, i.e., residual networks \cite{he2016deep, zhang2017residual,han2017deep,yu2017dilated,behrmann2019invertible}, originally designed for hardware or algorithmic efficiency, also exhibit physiological and functional parallels \cite{liao2016bridging}, prompting numerous comparisons with humans \cite{fleuret2011comparing,fogel2006evolutionary,kasabov2008evolving}.}

{In this work, we begin by putting emphasis on brain-inspired models with structural/functional connectivity, development, and learning mechanisms and their translations to the computational domain. Secondly, machine models can collaborate (e.g., via Mixture-of-Experts (MoE)), forming human-assisted systems in which functional complementarity can be leveraged.} Finally, models can {adapt} distribute the emergence of independent intelligences, exploring novel ways of task completions and implementation techniques.

\subsection{ Comparisons between HIns, HAss and HInd Intelligences}

%The distinction between "human-inspired" and "human-assisted" AI lies in the degree of involvement and influence that humans have in the development and operation of artificial intelligence systems. A summary of this taxonomy is illustrated in Fig. \ref{fig:intelX} with explanations for the time course of each of these approaches. 

%The distinction between ``human-inspired" and ``human-assisted" pertains to the extent of human involvement and influence in the development and operationalization of artificial intelligence systems. 
{The distinction between HIns and HAss lies in the degree of human involvement in developing and operating AI systems.} A summary of this taxonomy is illustrated geometrically in Fig. \ref{fig:intelX}, along with the temporal progression (directed arrows) of each paradigm in the evolutionary trajectory of the work that has been done in the last few decades. 
More specifically, assuming stable centroids of the studies conducted around human and machine intelligences, a \textbf{\textit{human-inspired}} (\textit{HIns}) trajectory will expand future works around these centroids, leading to more commonalities, i.e., increased biological plausibility due to everlasting incremental research around similar ideas. In the \textbf{\textit{human-assisted}} (\textit{HIns}) approach, the trend is reverse, i.e., the future overlap between machine and human intelligences tends to shrink, allowing hybrid designs to emerge and have a better capacity to effectively address the intelligence problem. Human assistance is not used as part of the training process but rather {treated} as the essential component of the system. On the contrary, \textbf{\textit{human-independent}} (\textit{HInd}) trend moves the centroids of the literature works in different directions while their expansion is likely to continue, potentially creating more overlaps between the two distinct intelligences {(also see Fig. \ref{fig:AHIplot})}. %Fig. \ref{fig:AHIplot} illustrates these options in detail and in the following subsections, we shall closely explore each one of these options along with the associated subcategories. 

\begin{figure*}[t!]
    \centering
    \includegraphics[width=0.8\columnwidth]{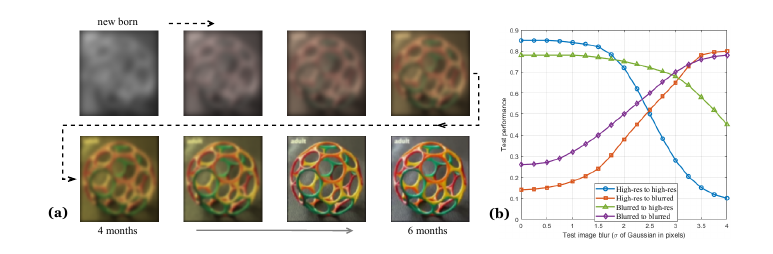}
    \caption{\textcolor{lightgray}{\textbf{(a)} An example of biomimetic progression of image resolution and color cues in the early months of development. Color maturization happens based on the wavelength of light i.e., red first accompanied by orange, followed by yellow, green, blue and purple towards the end of the spectrum. \textbf{(b)} Impact of training paradigms on performance with test images degraded by varying blur levels (quantified as the Gaussian kernel’s $\sigma$ in pixels): the blurred-to-high-resolution paradigm yields the strongest performance and generalization \cite{vogelsang2018potential}.}}
    \label{fig:biomimetic}
\end{figure*}

\subsubsection{HIns Intelligence
(Neuroscience meets Machines)}

{In HIns AI, machines are designed to mimic or replicate human intelligence, behavior, and cognitive processes. This endeavour draws on brain mechanisms for information processing and attentional control in complex tasks. Its aim is to understand and replicate the neural mechanisms underlying learning, common-sense reasoning, perception, and decision-making.}
%In Human-inspired AI, machines are designed to mimic/replicate aspects of human intelligence, behavior, or complex cognitive processes. Part of this endeavour could be inspired by the brain mechanisms responsible for processing information and navigating attention to complete complex tasks. Human-inspired AI aims at understanding and replicating the neural mechanisms behind human cognitive processes, such as learning, common sense reasoning, perception, and critical decision-making. %This approach often involves studying biological systems closely, including neuroscience, psychology, cellular chemistry, and cognitive sciences to accurately inform and guide the design of AI algorithms and architectures. 
The most critical phase of the human--like intelligence formation is the brain development and its training with  psychophysical limitations impacting its evolution. Examples include neural networks inspired by the brain's architecture (structural and functional connectivity) and specialization of the specific human brain regions, biologically plausible learning algorithms \cite{bengio2015towards} or algorithms (recursive, top-down and bottom-up) that accurately model human decision-making processes. %Human-inspired AI can be designed through methodological changes in model architecture (neural proliferation and pruning) or learning. %Architectural changes could be implemented through structural or functional changes (connectivity/topology) involving the addition (proliferation) or removal (pruning) of neural units in the training/learning processes. 

Machines can be inspired by humans in various conspicuous ways. For instance, humans are shown to perform blur-detection tasks using two different dominant strategies \cite{arslan2023distance}, {that} are {shaped by} various factors including top-down processes and behavioral similarities/differences. Behavioral characterizations of such human proclivities \cite{vssMatt, vssMichal} can be quantified and transferred to machine implementations to advance their guessing strength for {forced} multiple-choice tasks. The guessing framework is a mathematically sound tool to design {and} evaluate neural network strategies {leveraging} optimal guessing techniques as part of learning \cite{GuessingEntropy,guessingArslan}. {Psychophysics and neural activations might help reveal the temporal order of key processes and the spatial distribution of perceptual functions. Such insights are crucial for selecting inductive biases, architectures, and learning schemes while incorporating human decision-making into models. Fortunately, the human brain can be region-specifically monitored using neuroimaging techniques such as EEG \cite{siuly2016eeg}, MEG \cite{darvas2004mapping}, fMRI \cite{bellec2006identification}, and fNIRS \cite{cao2021brain}. These techniques enhance our understanding of the brain’s modular organization and the causal order of temporal/spatial processing. However, no equivalent tools exist to measure similar features in deep networks (e.g., modularity), despite the recent attempts based on information and statistical learning theories \cite{ hintze2023detecting}.} %ziv2005information

{One of the most remarkable stages of human brain development is the postnatal period, encompassing the first few years of life when new neural connections emerge at astonishingly rapid rates.} This dynamic mission of architectural modifications can be shown to be beneficial even in normal brain development \cite{paolicelli2011synaptic}. In the meantime, frontend optics as well as retinal anatomy also develop and start allowing more coherent light to enter and more cells to participate in retinal image processing, respectively, for more precise spiking, higher-level and accurate perception. It is shown in several past studies that poor visual acuity in human newborns offers interesting advantages, including faster and more accurate acquisition of visual object categories at multiple hierarchical levels \cite{Jinsi2022.06.22.497205}. The structural changes in every part of {the} visual system, such as chromatic and contrast sensitivities and its impact on the backend cortical substrate and functional connectivity, can be used to design biologically plausible (biomimetic) training regimens to help train the machines in a similar fashion \cite{Jinsi2022.06.22.497205,shah2022less,vogelsang2018potential}. %For instance, a training regimen that begins with high blur (low spatial resolution), gradually reducing its effect: allowing more high spatial frequency to gradually enter, mimics human vision development, and is shown to enhance basic-level categorization performance compared to using full resolution images throughout the training. %low2006axon
{For example, a training regimen that starts with high blur and gradually increases spatial resolution—mimicking human visual development—has been shown (e.g. as shown in Fig. \ref{fig:biomimetic}.b) to improve basic-level categorization compared to training with full-resolution images throughout \cite{vogelsang2018potential}.}
These progressions may hold adaptive value by fostering the formation of receptive field structures that confer resilience to subsequent spatial or chromatic degradations.  Fig. \ref{fig:biomimetic} demonstrates an example of biomimetic progression of acuity and color, applied to a toy image. As can be seen while the initial days of the postnatal period see gray scale, the maturation of cones leads to the processing of colors later in development. {In vision development, spatial frequencies and color change dynamically, alongside anatomical structure, internally generated noise, and maturation of recurrent pathways. Together, these factors refine input–feedback–consolidated architectures over time, particularly during infancy \cite{keunen2017emergence, wierenga2016development,setton2023age}.} %However, in vision development, not only are spatial frequencies and color subject to dynamic changes, but also the anatomical structure, generated internal noise over time, and maturation of recurrent pathways contributing to this biomimetic progression narrative. These factors collectively perfect the input feedback-consolidated architectural refinements over time, more so in infancy \cite{keunen2017emergence, wierenga2016development,setton2023age}.

One of the well-known barriers to good generalization is ``overfitting". To practically combat against it, regularization is proposed by injecting noise into the system to ensure bare minimum robustness \cite{noh2017regularizing, zhou2019toward} or expedite the training time \cite{audhkhasi2016noise}. The addition of noise into training is quite similar in spirit to noisy optimizations in \textit{simulated annealing} \cite{gutjahr1996simulated}, %jeffrey1986optimization
{helping models escape local minima by exploring a broader loss landscape.} By introducing noise, the optimization process can avoid getting trapped in small basins and suboptimal points, thereby improving the chances of finding a potential global minimum located inside a large basin. This mechanism is particularly useful in deep neural networks, where the loss landscape is highly non-convex with many local minima \cite{li2018visualizing}. 
%Understanding the learning mechanisms in humans, particularly the effects of attention and focus on the learning process, is crucial. These biological phenomena parallel computational techniques that mitigate erroneous learning and facilitate fact-checking, thereby ensuring the accuracy of information processed by the belief subsystem.

%Whether developmental changes in neural noise adhere to a similar rationale researchers have found in acuity, contrast sensitivity and color changes is still open. 
{Whether developmental changes in neural noise follow a rationale comparable to that observed in acuity, contrast sensitivity, and color perception remains an open question.} An attempt has been made recently to explore the interplay of simultaneous progressions of acuity, color and noise \cite{suaybCCN24}. These {studies} suggest that it is not unreasonable to expect that the temporal progression in neural noise could contribute to improved categorization performance, particularly under challenging environmental conditions. {A key question is how the brain leverages computational mechanisms, like generating internal noise, to enhance performance. Unlike linear systems that fail to capture complex behaviors and degrade with noise, non-linear systems such as the human brain exhibit multistability and can explore diverse network configurations when sufficient internal noise exists \cite{mcintosh2010development}.} %The more intriguing question is the computational mechanisms that the brain puts in place, such as generating the supply of noisy transmissions to realize this performance improvement. Unlike linear systems, which struggle to model complex behavioral dynamics and suffer from reduced information loss in the presence of noise, non-linear systems, {such as human} brain, can express multi-stability and explore various functional network configurations, provided sufficient internal noise is present \cite{mcintosh2010development}. %Internal noise is an intrinsic part of neural dynamics, yet in many practical scenarios, it is also influenced by external stimulation and visual (perceptable) noise. 
{Internal noise is an inherent aspect of neural dynamics, yet in many practical contexts it is also modulated by external stimulation and perceptible visual noise.} This interaction is significant due to the emergent behaviors, commonly referred to as \textit{stochastic resonance} (SR) in biological systems. SR manifests when a nonlinear system's signal-to-noise ratio is enhanced at moderate noise levels. At these noise levels, the signal is able to reach the threshold without being overwhelmed by the noise, thereby optimizing signal transmission \cite{gammaitoni1998stochastic}. As a result, the interplay between non--linearity and internal noise becomes crucial for the spontaneous emergence of exploratory dynamics, even without external stimulation or noise. %Observation of such biological phenomenon could be extremely useful and inspire us to design and train computational models that would have human--like behavioral dynamics in resolving complex tasks. 

Another source of inspiration could relate to the specialized brain pathways to process distinct spatial frequencies for enhanced task completions under severe environmental conditions. Early studies on spatial frequency disentanglement in the brain suggest that low and high frequencies are processed distinctly, involving separate visual pathways that develop later in life to enhance human robustness in identification and categorization tasks \cite{kauffmann2014neural}. {To translate human studies into ML, adjust training regimens -- e.g., include task-relevant data or image degradations -- to align training and test distributions. In contrast, biological systems evolve through continual environmental interaction, where sensory-driven plasticity and co-activation of neurons (Hebbian wiring) produce specialized, strongly connected circuits.} %The most straightforward translation of human studies to machine model development or learning involved altering training regimens to include statistically relevant data or introducing various degradations in the image space to align the underlying distributions driving training and testing  conditions. However, biological processes follow a path coupled with the environment where external stimuli directly {impact} the formation of new neuronal pathways, resulting in the neural connection changes and the formation of specialized neural populations that fire together. Simultaneous firing eventually forms strongly connected networks (wiring in Hebbian sense) in specialized brain regions. 

{The HIns paradigm provides a crucial path to mitigate the risks of algorithmic misuse and unintended harm. By focusing on bio-plausibility (e.g., modularity and developmental learning), HIns systems are designed to be more interpretable and aligned with human cognition, which is essential for accountability and auditing in critical domains. Ultimately, the HIns goal is to organically embed human values like fairness into the design, ensuring responsible and ethical advancement, rather than guaranteeing improved accuracy.}

\subsubsection{HAss Intelligence (Birth of Hybrid Systems)}

In HAss AI, models are developed with the primary goal of augmenting or enhancing human/machine capabilities rather than replicating them. HAss leverages AI technologies to assist machines or humans in performing tasks more efficiently, accurately, or autonomously. {HAss, as a bidirectional enterprise, can be described as the ongoing interaction between humans helping machines and machines helping humans. However, fully realizing the potential of Human–AI collaboration requires overcoming key challenges. First, establishing the conditions for complementarity—where HAss performance exceeds AI alone—depends on humans recognizing when to augment AI in critical decision-making. Second, accurately assessing human mental models and susceptibility to error is essential, highlighting the impact of design choices in human–AI interaction, such as the timing and extent of human assistance to mitigate bias.} %However, fully realizing the potential of Human--AI collaboration requires addressing important challenges. First, the necessary conditions that support complementarity -- where HAss performance is expected to surpass that of AI alone for a given task, requiring humans to recognize when to augment AI that complements critical decision--making processes. Second, accurately assessing human mental models as well as the extent to which humans are prone to failures is crucial, which also brings out the issue of understanding the effects of design choices in human--AI interaction, such as the timing and amount of human assistance to avoid bias. 

\footnotesize
\begin{table*}[t!]
  \centering
  % Set alternating row colors
  \rowcolors{2}{gray!10}{white}
  \begin{tabular}{
      >{\raggedright\arraybackslash}p{1.3cm} 
      %L{1.4cm}
      >{\raggedright\arraybackslash}p{2.2cm} 
      %L{2.2cm}
      >{\raggedright\arraybackslash}p{4cm} 
      >{\raggedright\arraybackslash}p{3.1cm} 
      >{\raggedright\arraybackslash}p{2.7cm} 
      >{\raggedright\arraybackslash}p{1.8cm}}
    % Header row with background color
    \rowcolor{blue!20}
    \textbf{Category} & \textbf{Goal} & \textbf{Design Ideas} & \textbf{Strengths} & \textbf{Weaknesses} & \textbf{References} \\
    \toprule
    \textbf{HIns} (Human-Inspired) & Mimic human cognitive processes and decision-making. & \vspace{-3mm}
    \begin{itemize}[leftmargin=*, noitemsep]
      \item Leverage neuroscience insights. 
      \item Model brain architecture.
      \item Emulate functional connectivity.
      \item Use developmental learning.
    \end{itemize} &
    \vspace{-3mm}
    \begin{itemize}[leftmargin=*, noitemsep]
      \item Enhanced interpretability.
      \item Closer alignment with human cognition.
      \item Potential for common-sense reasoning.
    \end{itemize} &
    \vspace{-3mm}
    \begin{itemize}[leftmargin=*, noitemsep]
      \item May inherit human biases.
      \item Complex to model accurately.
      \item Limited scalability.
    \end{itemize} &
    \vspace{-3mm}
    \begin{itemize}[leftmargin=*, noitemsep]
      \item \hspace{-1mm} \cite{liao2016bridging}, \cite{petrini2014vision}, \cite{bengio2015towards}, \cite{arslan2023distance}, \cite{vssMatt}, \cite{vssMichal}, \cite{paolicelli2011synaptic}
    \end{itemize}  \\
    \midrule
    \textbf{HAss} (Human-Assisted) & Augment human or machine performance through collaboration. &
    \vspace{-3mm}
    \begin{itemize}[leftmargin=*, noitemsep]
      \item Integrate human input.
      \item Combine with automated processes.
      \item Leverage interactive LLMs.
      \item Employ decision-support tools.
    \end{itemize} &
    \vspace{-3mm}
    \begin{itemize}[leftmargin=*, noitemsep]
      \item Merges human intuition with computational efficiency.
      \item Improves performance on complex tasks.
    \end{itemize} &
    \vspace{-3mm}
    \begin{itemize}[leftmargin=*, noitemsep]
      \item Dependent on human involvement.
      \item Integration can be complex.
      \item May introduce variability.
    \end{itemize} &
    \vspace{-3mm}
    \begin{itemize}[leftmargin=*, noitemsep]
      \item{\hspace{-1mm} \cite{shazeer2017outrageously}, \cite{liu2024deepseek}, \cite{brown2020language}, \cite{touvron2023llama}, \cite{shankar2024validates}, \cite{kalyanpur2024llm}, \cite{makhataeva2023augmented},  \cite{kaur2024ai}}
    \end{itemize} \\
    \midrule
    \textbf{HInd} (Human-Independent) & Create autonomous AI systems operating without direct human mimicry. &
    \vspace{-3mm}
    \begin{itemize}[leftmargin=*, noitemsep]
      \item Focus on algorithmic efficiency.
      \item Optimize for hardware performance.
      \item Utilize back--propagation, CNNs, and transformer architectures.
    \end{itemize} &
    \vspace{-3mm}
    \begin{itemize}[leftmargin=*, noitemsep]
      \item High efficiency and scalability.
      \item Less influenced by human biases.
    \end{itemize} &
    \vspace{-3mm}
    \begin{itemize}[leftmargin=*, noitemsep]
      \item Lacks human-like interpretability.
      \item May struggle with nuanced tasks.
    \end{itemize} &
    \vspace{-3mm}
    \begin{itemize}[leftmargin=*, noitemsep]
      \item \hspace{-1mm}  \cite{krizhevsky2017imagenet}, \cite{hintze2023detecting}, \cite{li2022fusion}, \cite{shamay2019real}, \cite{francken2018neuroscience}, \cite{krugliak2022towards}, \cite{hagendorff202015}, \cite{shoeybi2019megatron},
      \cite{bommasani2021opportunities}
    \end{itemize} \\
    \bottomrule
  \end{tabular}
  \caption{\textcolor{lightgray}{Comparisons between Human-Inspired (HIns), Human-Assisted (HAss), and Human-Independent (HInd) Intelligences and corresponding reference articles that fall in these categories.}}
  \label{tab:comparison_fancy}
\end{table*}
\normalsize

Primary example of HAss is the foundation models (FMs) that heavily interact with humans, such as most recent advanced LLMs \cite{brown2020language} (GPT4 \cite{bubeck2023sparks}, Llama \cite{touvron2023llama}, etc.) that begin transforming the software industry by providing assistance writing the bulk of the source code as a downstream task, and still subject to thorough validation \cite{shankar2024validates, kalyanpur2024llm}. {A critical enabler of this human–AI synergy is reinforcement or task learning from human feedback (RLHF/TLHF), where human evaluators guide model optimization through a reward mechanism trained on raters' ranking to ensure outputs remain aligned with human preferences, safety constraints, and contextual requirements.} This approach emphasizes the significance of collaboration between humans and machines, with AI systems providing support, guidance, or automation to complement human skills (critical thinking, deep reasoning, navigation of complex decisions) and expertise. Other examples include AI--augmented tools for object recognition, memory enhancement \cite{makhataeva2023augmented}, data analysis, medical diagnosis \cite{kaur2024ai}, language translation, and robotic assistance in manufacturing or logistics \cite{zeng2023large}. 
%Yet another hot topic could be mission critical applications where humans and machines could perform tasks using different strategies and hence end up having distinct failure patterns. This complementarity can be utilized as a fusion to bolster overall accuracy performance \cite{groen2017hybrid}. 

{Human cognitive and sensorimotor organizations offer valuable inspiration for computational model developments. A notable example is eye tracking, where gaze direction provides insights into internal cognitive states. Gaze has been shown to reveal attentional patterns \cite{posner1990attention}. More generally, gaze data is useful in two ways: (i) it can guide the design of AI attention mechanisms to identify salient information in complex environments, potentially reducing the data requirements for learning adaptive attention by leveraging human gaze directly \cite{mishra2017learning}; and (ii) it can enable AI systems to interpret gaze for naturalistic, multimodal human–machine interaction, eliminating the need for traditional input devices and supporting more immersive interfaces \cite{li2022fusion}.}

{HAss systems seek to achieve synergy through the efficiency of machines to augment human cognitive and physical capabilities, rather than replace them. This human-centered strategy recasts the nature of labor, shifting the economic discussion from job replacement to the creation of new, higher-paid hybrid human-machine jobs. HAss tools make this possible through automated routine, which allows human experts to focus on tough decisions, ethical dilemmas, and creative problem-solving, thereby evading the systemic risk of technological unemployment.}

\subsubsection{HInd Intelligence (Algorithms for Machine Hardware)}

In our taxonomy, this branch is still considered as human-centered AI because the studies in this category are still framed based on human intelligence as a reference model. For instance, although AlexNet \cite{krizhevsky2017imagenet} is inspired by the visual cortex and its layered functional structure, designed to mimic human visual perception and processing, the proposed optimization algorithm  Back-Propagation (BP)  is not necessarily human-inspired; it is more inspired by how computational constraints shape and efficiently utilize resources, along with mathematical convenience such as sticking to the same set of weights/coefficients in both forward and backward passes \cite{xiao2018biologically}, {and yet works exceptionally well in practice}. In fact, BP is an optimization method which spreads information across all parameters of the neural network making the network less modular \cite{hintze2023detecting}. Hence the development of AlexNet is not necessarily categorized as human-inspired but rather \textit{neuroscience-inspired} at the very high levels of abstraction subject to SoTA silicon hardware constraints and explainability.

%The main motivation behind HInd is twofold: \textbf{(1)} The pursuit of an open intelligence could potentially be a better model for the natural world we live in as human intelligence seems to have missing elements to tackle important issues of the world such as less advanced predictive capabilities for extended time and range or the inherent limitations of convolution/pooling operations executed by the visual system. This is corroborated by the most recent transformer models which lift off some of the important inductive biases that previous generation CNNs adopted. 
{The main motivation behind HInd is twofold: \textbf{(1)} The pursuit of open intelligence may provide a better model of the natural world, as human intelligence shows limitations in addressing global challenges—for instance, weaker predictive capabilities over extended scales or constraints of convolution/pooling operations in the visual system. Recent transformer models support this view by overcoming key inductive biases (texture etc.) inherent in earlier CNNs.} Another observation is the ever--growing data scales, as machines having less inductive bias would require to build such biases automatically based on the peculiarities of voluminous data. \textbf{(2)} In any reliable/efficient AI system,  we would expect a joint symbiotic existence between the intelligence and the hardware that runs it. The common Von-Neuman architecture is not necessarily the right home for the optimal implementation of human intelligence. This might guide us to think of other forms of intelligences that could optimally run on the available hardware architectures, building on the few-decade experience of distributed microprocessor history of human construct \ref{VNeffectonAI}.

{It may be more accurate to frame HInd as an approach that regards humans as largely irrelevant to large-scale AI development. A growing view in the community suggests that independent design can exploit insights from real-world interactions and experiments, which often yield counterintuitive results not captured by intrinsic data. Nevertheless, such systems remain constrained by human-defined metrics and are frequently benchmarked against human performance to enhance explainability.} After all, explainability is a human construct and only meaningful if it helps with understanding the true nature of the phenomenon.  {HIns architectures (e.g., spiking networks, local learning rules) often improve interpretability and sample efficiency but impose slower training, limited accelerator parallelism, greater optimization complexity, and immature toolchains. HInd models (e.g., CNNs/Transformers trained with BP) scale efficiently on modern hardware yet are less interpretable and typically demand substantially more data and energy. Discussions of biological plausibility should therefore report trade-offs in runtime, memory, energy and parallelizability.} In Table \ref{tab:comparison_fancy}, we present a summary of the \textit{HIns}, \textit{HAss}, and \textit{HInd} intelligences and the  corresponding studies for streamlined reference and comparative analysis.

%Perhaps it might be more accurate to frame HInd as an effort which characterizes humans as irrelevant to the next generation large--scale AI development. There is a growing inclination in the community believing that the independent design will benefit from some of the findings that come out through interactions with the outside world (experiments etc.) which might well be counter-intuitive and cannot be adequately represented by the data intrinsics. However, these designs are still bound to subjective evaluations by human--credited metrics and might be compared to human performance to improve their explainability. After all, explainability is a human construct and only meaningful if it helps with understanding the true nature of the phenomenon. 

\begin{figure}[b!]
    \centering
    \includegraphics[width=0.7\columnwidth]{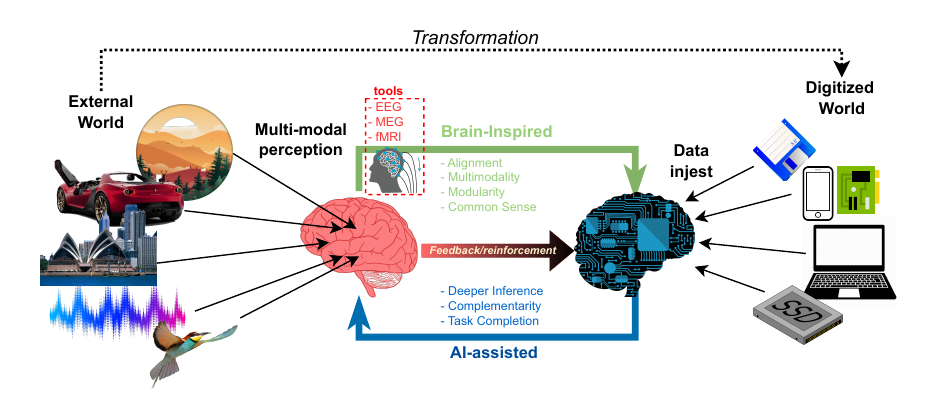}
    \caption{\textcolor{lightgray}{Elements of Brain-inspired information processing.}}
    \label{fig:biip}
\end{figure}

\section{Human-Level AI and Challenges/Perspectives}\label{section5}

%By combining insights from human cognition (human-inspired) and biological neural structures (brain-inspired), human-level AI strives to achieve general intelligence while maintaining ethical, contextual, and adaptive capabilities similar to humans.

\subsection{Brain-inspired Information processing }
\label{section4}

{Brain-Inspired information processing, subsumed by human-inspired processing, at the top of the intelligence inspiration tree (Fig. \ref{fig:IntelTypes}), involves designing computational systems and architectures that emulate the brain’s structure and function to achieve advanced cognitive and data-processing abilities. This interdisciplinary field draws on neuroscience, cognitive science, and AI to develop models for tasks such as multi-modal perception, developmental learning, and critical decision-making. Fig. \ref{fig:biip} illustrates the key elements and context of brain-inspired information processing \cite{zhao2023brain}. Neuroimaging techniques, measuring electrical and magnetic activity, provide high temporal resolution to track dynamic neural activity \cite{attal2009modelling}, revealing the neural dynamics and connections underlying complex cognition. Machine learning algorithms can identify neural signatures for specific functions, while statistical methods, such as random graph theory, model brain networks and highlight key regions involved in information processing \cite{bullmore2009complex}.}

The informative features extracted from the biomedical data are typically used to build computational models that mimic the way our brains process information. Translating these measurements into computational frameworks that are truly brain-inspired could be quite challenging.  For example, neural networks can be designed to reflect the hierarchical nature of the brain, with different layers corresponding to various levels of processing found in the visual cortex \cite{kriegeskorte2015deep}.  %zhao2023brains %Finally, these computational frameworks are refined by comparing them against real brain data, perhaps at higher abstraction levels, ensuring they not only mimic brain activity but also can generalize to new, unseen tasks. However,  the resemblance is always subject to questions as it is hard to completely translate brain operations to algorithms and implement them on silicon. 
{Finally, these computational frameworks are refined by comparison with real brain data, often at higher abstraction levels, ensuring they not only mimic brain activity but also generalize to unseen tasks. However, the resemblance remains questionable, as it is difficult to fully translate brain operations into algorithms and implement them on silicon.} By blending neuroimaging with AI and statistical analysis, researchers mainly aim to create brain-inspired models that capture, at least to some extent, the core manifestation of human-level cognition. These models can then be applied to fields ranging from the most popular FMs to autonomous systems. In fact, brain-inspired data processing could guide FMs to embrace emergent behaviours more rapidly as architectural similarity is more likely to demonstrate functional similarities \cite{haber2022learning} (See also Section \ref{concSection} for further discussions). 
%This approach seeks to bridge the gap between biological intelligence and machine intelligence, pushing toward more flexible and robust AI systems. 

\subsubsection{Reverse engineering a human skill} {A core aim of brain-inspired research is to understand the emergent behaviors underlying intelligence. Prior work suggests that the difficulty of reverse-engineering a human skill correlates with how long it has evolved in living beings \cite{bharath2008next}. For instance, perception—one of the oldest and seemingly effortless skills—is hypothesized to be among the hardest to replicate, whereas effortful skills may be comparatively easier to engineer.}

%One of the main pillars of brain-inspired research is to understand how hard it is to fully grasp the various emergent behaviors that lead to intelligence in the brain. %This question dates back to the 1980s and the difficulty of adapting it depends on the developmental trajectory of that skill.  The premise of the past work was that we typically expect the difficulty of reverse-engineering any human skill to be roughly proportional to the amount of time that skill has been evolving in living beings, including animals \cite{bharath2008next}. For example, one of the oldest human skills is perception, which is largely unconscious and appears to be effortless. The hypothesis is that we should expect skills that appear effortless to be difficult to reverse-engineer. The dual to this hypothesis is that skills that require effort may not necessarily be difficult to engineer at all.

Emerging skills developed later in the evolutionary trajectory, such as mathematics, logic, and scientific reasoning, pose challenges as they diverge from our evolutionary predispositions. These abilities have developed relatively recently in historical terms, undergoing refinement mainly through cultural evolution over the past few thousand years. Initially, many prominent researchers assumed that once they had grappled with the ``hard" problems, the seemingly ``easy" challenges of vision and commonsense reasoning would fall in place. However, the last decade in AI has proven it wrong. {This misconception arose from recognizing that seemingly simple problems are in fact highly complex, rendering machines’ proficiency in tasks such as logic and algebra largely irrelevant, as these are trivial for them to perform \cite{brooks2018intelligence}.} %zhang2023toward
%This misconception stemmed from the realization that these supposedly straightforward problems are, in fact, incredibly difficult, leading to the machines' proficiency in solving problems like logic and algebra being irrelevant, as these tasks are exceptionally easy for them to handle \cite{zhang2023toward,brooks2018intelligence}. 

%Human memory is a complex, multi-layered system that has evolved primarily to support survival-driven tasks such as pattern recognition, spatial navigation, and social interactions \cite{squire2011cognitive}. Unlike artificial memory systems, which store information with high fidelity, human memory is associative, reconstructive, and prone to biases. To reverse engineer it, one must understand its fundamental mechanisms—encoding, storage, and retrieval—along with its interactions with attention and emotion. Recent advancements in Multi-Head Latent Attention (MLA) \cite{meng2025transmla} demonstrated how low--rank compression of Key--Value (KV) matrices mimics human memory’s selective retrieval and reconstruction of sparse cues. For instance, \textbf{DeepSeek}’s MLA reduces KV cache sizes by 16$\times$, akin to episodic memory’s efficiency in reconstructing details from minimal sensory input \cite{moscovitch2016episodic}. This aligns with neuroscientific observations of how humans compress and recall information without exhaustive storage, a principle also adapted in long-context AI tasks like document analysis \cite{kwak2022unveiling}.

\subsubsection{What makes humans special and/or not so special?} {Humans show both strengths and limitations in recognition, perception, detection, and search. They are not adept at detection tasks like blur identification or misaligned lines, with performance degrading sharply without focused attention. They also struggle with hidden correlation tasks, as estimating joint probability distributions across many random events exceeds cognitive capacity and the demands of integrating multi-modal imprecise information \cite{gopnik1999scientist}.} %Humans exhibit unique strengths and limitations in tasks like recognition, perception, detection, and search. First and foremost, humans are not adept at ``detection" tasks like blur identification or identifying misaligned lines. Especially without focused attention, their performance significantly degrades, making it difficult to excel in completing such tasks successfully. Additionally, humans struggle with "hidden correlation" tasks, as their ability to estimate the joint probability distribution of a large set of random events is very limited \cite{gopnik1999scientist}.  This limitation arises from the cognitive load and complexity involved in processing and integrating multi-modal sources of imprecise information simultaneously. 

However, human cognitive system is extremely good at visual search \cite{eckstein2011visual} and finding objects/entities (i.e., good at ontologies -- a set of concepts and categories that shows their properties and the pairwise relations) leading to good performance in languages. Humans also have a sparse but adequate understanding of physical reality and causality, which leads to data processing efficiencies in their generalization capacities, {forming} learned world-models to be usually reasonable and in congruence with the required tasks. They are also good at grounding the language with mental simulations. {Human perception is largely organized around objects—primarily within three dimensions. These objects serve as the building blocks of higher cognition, supporting language, planning, and reasoning. Conceiving the world as separable parts that can be independently processed and recombined allows humans to generalize far beyond direct experience.} %It is widely recognized that the way humans perceive the world is organized around objects - mostly up to the third dimension and some lousy understanding of forth dimension. These objects act as fundamental building blocks for a variety of higher-level cognitive processes, including language, planning, and reasoning. The notion of the world being composed of separate parts that can be processed autonomously and combined in countless ways enables humans to make generalizations that go well beyond their direct experiences. 

%pearl2018book 
Let us consider our question with respect to the three rungs of the ladder {of causation} \cite{moruzzi2022climbing}: \textbf{(1)} The initial stage involves the extraction of meaningful correlations from real-world representations through observation. This endeavor aims to estimate the conditional expectations of events, thereby {forming} crucial associations. \textbf{(2)} The subsequent stage involves engaging with the environment using an accurate world model that might be formed partially in the previous rung. Interactions in this phase result in alterations to the characteristics of objects that are inherently linked or correlated, either directly or through other dependent or confounding variables. Interventions, occurring over time, engender a sequence of actions and reactions that are conducive to causal reasoning. \textbf{(3)} The final stage of the ladder, termed \textit{counterfactual thinking}, necessitates envisioning alternative scenarios, evaluating them, and adjusting accordingly. {This aspect requires imagining changes and their repercussions without active intervention, highlighting counterfactuality as one of humanity's most demanding cognitive abilities: predictive reasoning.} %This aspect is distinguished by its requirement for imagining changes and their potential repercussions, without necessarily engaging in active intervention. Achieving counterfactuality underscores one of humanity's most demanding cognitive abilities: predictive reasoning.

\subsubsection{Modularity and Robustness}
\label{SectionMod} %The brain network modularity can be defined as the degree to which functional brain networks (connectivity) are divided into special subnetworks. 
Modularity is assumed to be beneficial to develop a robust brain against internal as well as external perturbations. Biological individual differences suggest the mandatory modular organizations for the robustness of persistent activity to various perturbations \cite{chen2021modularity}. Modularity does not typically form in computational models with standard training and architectures. However, modularity can be enforced through various approaches such as changing the training or the optimization process. For instance, biomimetic training is shown to demonstrate more robust behavior under out-of-distribution scenarios {(see Fig. \ref{fig:biomimetic}.b)}. In addition,  biologically plausible optimizations provide robustness and seem to allow the emergence of invariant representations more quickly \cite{stanley2019designing}.

One way to achieve robustness in recognition tasks is to retain information in a redundant and distributed manner. A widely accepted assumption is that distributed information leads to robustness to internal and external perturbations, whereby lost or corrupt information in one brain region can be compensated for by redundant information in other regions \cite{barrett2016optimal}. Redundant modular representations might naturally emerge in neural network models that could learn robust dynamics \cite{chen2021modularity} as long as such representations can be retained in later stages of training. Indeed, developing modularity might allow learned features to be frozen and the rest of the other hardware could be used to learn new tasks allowing learning rates to change over time for different parameters of the neural system \cite{kirkpatrick2017overcoming}. {Modularity can enhance robustness to external perturbations, though robust systems do not always exhibit modular organization.} %Modularity has the potential to lead a system to be more robust to external perturbations. However, not every robust system necessarily manifests any sort of modularity in its operational characteristics.

Having observed biologically inherent modularity naturally encourages researchers to investigate the architectural changes required for machines to acquire human-like robustness, which can be provided by, for instance, the brain's functional modularity. For instance, MoE architectures \cite{shazeer2017outrageously} {coupled with Multi-head Latent Attention (MLA) \cite{vaswani2017attention} or variants (GQA)}, used in DeepSeek \cite{liu2024deepseek} {or Llama-4 \cite{meta2025llama4}}, could replace dense feed-forward layers with dynamic expert routing, {forming inherent modularity and} activating specialized sub-networks per input. {The combination of MLA/GQA and MoE forms an efficient architecture by tackling memory and computation bottlenecks in LLMs. MoE uses sparse activation, routing tokens to a few specialized “experts”, enabling scaling to trillions of parameters at low inference cost. MLA compresses the Key-Value (KV) cache via low-rank factorization, allowing very long context windows with minimal memory. Together, MLA’s memory savings support more MoE experts. Benchmark results are given in Table \ref{tab:mla_benchmarks}.}

%This mirrors the brain’s modular specialization, where distinct neural circuits handle specific tasks (e.g., visual vs. auditory processing). While not direct analogs, MoE systems might address challenges like load balancing and training stability, akin to how biological systems manage resource allocation across functional modules.

\begin{table*}[t]
    \centering 
    %\scriptsize 
    \footnotesize 
    \caption{Comparison of MLA-based models v.s. conventional architectures on major language benchmarks (MMLU, DROP, IF-Eval, GPQA) and efficiency metrics. The information is gathered from technical reports\cite{liu2024deepseek,meta2025llama4}, HuggingFace, etc.}
    \label{tab:mla_benchmarks}
    \begin{tabular}{|>{\columncolor{gray!15}}l|c|c|c|c|c|c|c|}
    \hline
    \rowcolor{blue!20}
    %\rowcolor{gray!15}
    \textbf{Model} & \textbf{Params} & \textbf{Atten.} & \textbf{MMLU} & \textbf{DROP} & \textbf{IF-Eval} & \textbf{GPQA} & \textbf{KV / Thr.} \\
    \hline
    DeepSeek-V2 (MoE) & 236B (21B active) & MLA & 78.2 & 83.0 & -- & 35.3 & -93.3\%, 5.76$\times$ \\
    \hline
    DeepSeek-V3 (MoE) & 671B (37B active) & MLA & \underline{\textbf{88.5}} & \underline{\textbf{91.6}} & 86.1 & 59.1 & -- \\
    \hline
    LLaMA-3.1-70B (Instr.) & 70B & GQA & 79.3 & 79.6 & \underline{\textbf{87.5}} & 46.7 & 0\%, -- \\
    \hline
    LLaMA-2-7B (TransMLA) & 7B & MLA (conv.) & $\approx$ orig.\ level & -- & -- & -- & -93\%, 10.6$\times$ \\
    \hline
    LLaMA-4 (MoE, Maverick) & 400B (17B active) & GQA & 85.5 & -- & -- & \underline{\textbf{73.5}} & -- \\
    \hline
    GPT-4o & $\approx 200$B & Dense & 87.2 & -- & 84.3 & 49.9 & 0\%, -- \\
    \hline
    Claude-3.5 Sonnet & $\approx 175$B & Dense & 88.3 & 87.1 & 86.5 & 65.0 & 0\%, -- \\
    \hline
    Mistral 7B & 7B & Dense & 60.1 (public) & -- & -- & -- & 0\%, -- \\
    \hline
    Mixtral 8$\times$7B & 47B (12B active) & MoE (dense att) & 70.6 (public) & -- & 63.1 & -- & -- \\
    \hline
    \end{tabular}
\end{table*}

\subsubsection{Missing elements towards achieving Human-level AI} \label{missingElements} {Without higher level abstractions, counterfactuality, predictive and commonsense reasoning and explicit compositionality, it is almost impossible to achieve or get to human--level AI in the near future. This corollary} summarizes the main elements missing in any design towards achieving human-level capabilities \cite{zhang2023toward,marcus2020next}.  {AI with commonsense should interpret everyday situations and make decisions using implicit knowledge humans take for granted. 
As noted in \cite{marcus2020next}, this entails building representations of core constructs such as time, space, causality, and modeling interactions between physical objects and humans.} %butz2021towards

Tools for few--shot generalization (without exponentially increasing the training data size) \cite{wang2020generalizing}, must be totally different than what {the} AI community is focusing on/and may require the next breakthrough in science and engineering. One approach to achieve such immediate generalization might be \textit{biological plausibility}, which is often exercised by the research community in diverse ways including the modifications of neural structure \cite{illing2019biologically}, inductive bias, training regimen \cite{marr2010vision} or activation patterns. % plausibility: rosa2013biologically
{In our context,} biological plausibility refers to how closely ANNs mimic the structure and function of biological neural systems, like spiking neurons in the human brain \cite{taherkhani2020review}. This involves replicating neural architecture, using local learning rules akin to Hebbian learning \cite{gerstner2002mathematical}, %or the advanced BCM theory \cite{bienenstock1982theory}
incorporating modification thresholds, learnable activation functions and dynamics, similar to biological neurons, achieving energy efficiency and providing a plausible explanation for synaptic scaling \cite{turrigiano2004homeostatic}. However, biological plausibility must be understood in the context of the scale at which learning is considered. For instance, if a deep network is viewed as an unrolled recurrence in the brain, an argument can be made in favor of its biological plausibility \cite{liao2016bridging}. Conversely, when examining fundamental operations such as convolutions and BP for parameter optimization, an argument can easily be made against their biological plausibility \cite{stork1989backpropagation}. %xiao2018biologically 

{Training spiking networks at scale remains an open engineering problem (surrogate gradients, non-differentiable updates), which limits their immediate applicability to very large problems despite their theoretical energy/latency benefits. It is also worth reconsidering the biological plausibility of core deep network operations such as convolutions, activation functions, and optimization rules. A key feature of convolutions in vision—weight sharing—is absent in biological networks \cite{grossberg1987competitive}. Attempts to relax this via locally connected networks generally underperform in large--scale architectures and high--class categorizations \cite{bartunov2018assessing, neyshabur2020towards}. Biologically plausible networks can enhance interpretability, robustness, adaptability, and energy efficiency. However, plausibility alone does not guarantee behavioral similarity nor does it ensure better performance, which depends on task complexity, generalization, and learning efficiency. Full behavioral mimicry requires broader considerations of intelligence and adaptability beyond structural and functional resemblance.}

%It is also {reasonable} to reevaluate the biological plausibility of basic operations of deep networks such as convolutions, activation functions or optimizations (cost function, learning rules) etc. One of the most critical aspects of convolutions used in computer vision is the weight sharing, a capability that biological networks do not possess \cite{grossberg1987competitive}. Several studies have attempted to relax weight sharing in convolutional networks by introducing locally connected networks \cite{bartunov2018assessing,neyshabur2020towards}. However, such networks perform worse for large--scale architectures and large--class categorizations.

%Networks that are biologically plausible can improve interpretability, robustness, adaptability, and energy efficiency, reflecting some of the advantageous properties of biological systems. However, biological plausibility does not necessarily ensure behavioral similarity, which is the resemblance in outputs and responses to those of biological systems. Achieving behavioral similarity requires addressing additional factors such as the complexity of tasks, generalization capabilities, and learning efficiency. While biologically plausible networks may show some behavioral similarities due to their design, achieving full behavioral mimicry involves broader considerations of intelligence, learning, and adaptability beyond just structural and functional resemblances.

Another missing item is the compositional understanding of symbol-like entities. %This encompasses the ability to recognize and represent objects not merely as isolated entities but as components of larger, more complex inter--related structures. In biological systems, this compositional understanding allows for the hierarchical processing of sensory information \cite{deco2017hierarchy}, enabling the perception of scenes, the understanding of contextual information, and the recognition of relationships between objects and their affordances. Incorporating this level of compositionality into ANNs could significantly enhance their ability to generalize from limited data, perform complex reasoning tasks, and interact more effectively with dynamic and multifaceted environments. 
{This encompasses the ability to recognize and represent objects not merely as isolated entities but as components of larger, inter-related structures. In biological systems, this compositional understanding allows hierarchical processing of sensory information \cite{deco2017hierarchy}, enabling scene perception, contextual understanding, and recognition of relationships between objects and their affordances. Incorporating this compositionality into ANNs could enhance their ability to generalize from limited data, perform complex reasoning, and interact more effectively with dynamic, multifaceted environments.} One particular computational manifestation of increased compositional understanding could be attributed to effective receptive fields of the stacked layers in CNNs \cite{luo2016understanding}, which could potentially be achieved either through using larger kernel sizes in convolutional layers \cite{ding2022scaling} or deeper architectures with smaller kernel sizes (e.g., ResNet etc.).  {Abstraction and compositionality must support reasoning under incomplete knowledge, while human-inspired learning systems continually integrate new information to foster deeper understanding {for} AGI. 
Structural parallels to human anatomy can further promote explainability and interpretability, enabling AI to justify its actions in ways comprehensible to humans. 
Such transparency is especially necessary  in mission-critical domains such as healthcare, where understanding AI-driven decisions can markedly be vital.}

{Another}  {Interpretation/implementation of learning in} {recent} neural network models is yet another barrier \cite{bommasani2021opportunities}. {While BP may not be biologically plausible,} the principles underlying BP can inspire the development of biologically plausible learning algorithms. Mechanisms such as \textit{feedback alignment} (FA), burst propagation, and predictive coding offer promising alternative pathways for approximating the learning capabilities in biological networks. The work by Lillicrap et al. \cite{lillicrap2020backpropagation} highlights the importance of these alternative mechanisms, bridging the gap between computational models and biology. 

{As AI systems increasingly assume mission-critical roles, ensuring their safety, security, and robustness against adversarial attacks, failures, and intrusions has become a central research priority. This involves methods for verifying and validating AI behavior and mitigating deployment risks. Some argue that missing elements may include self-awareness and self-assessment. Yet, the conceptualization of consciousness in AI remains one of the most elusive goals \cite{lewis2011survey}. Whether artificial entities could ever possess subjective experiences, such as a sense of safety, is still unknown. Attempts at mathematical modeling \cite{kleiner2020mathematical} have so far failed to account for the full spectrum of human conscious experience.}
%chella2020developing, namazi2012mathematical

\subsection{Challenges and Obstacles}

AI systems {are expected to} hold up ethical principles and societal values as part of their design and ensure that algorithms adopt fair, transparent, and unbiased behavioral characteristics. This undertaking involves navigating issues such as privacy, accountability, and the long-term impact of AI on biological intelligence, employment and other dynamics of society at large. {It is clear} that the next generation AI and its implementation will generate significant societal impacts, necessitating moderation, governance and legislative regulations. 

Training FMs and enhancing their ability to generalize knowledge across different domains/tasks through specialized fine tunings (transfer learning) is another major challenge. This adaptation to novel circumstances involves enabling FMs to learn novel concepts from only a few examples (a.k.a. few-shot learning \cite{wang2020generalizing}), leveraging previously acquired knowledge to solve new problems more efficiently. {However, as with modeling many cognitive faculties of the brain for real-world problem solving, no widely accepted mathematical model of transfer learning links out-of-distribution generalization to the minimum number of parameters needed for updates.} %However, as is the case with modeling many cognitive faculties of the brain with regards to real-world problem solving, there is no widely accepted mathematical model of transfer learning that connects out-of-distribution generalization to the minimum number of model parameters required for updates. 
Accomplishing such a goal involves many genuine techniques that would make machines look like humans. Of such mechanisms, meta-learning \cite{vettoruzzo2024advances} and self-supervised learning \cite{gui2024survey} are the two other critical areas of {HIns} AI research. Developing AI systems that can learn to learn, adapt, and improve over time through meta-learning techniques are desirable properties of the future generations of AI technology. 
%The overall goal of learning  involves designing algorithms that can automatically discover effective  strategies and apply them to new tasks, allowing AI to become more autonomous, capable and self-directed over time. 

{Integration} with physical bodies (embodiment) and sensory capabilities to enable interaction with the physical world is another significant challenge. The ``embodied intelligence" \cite{gupta2021embodied} involves understanding how intelligent behavior emerges from the complex interplay between an agent's hardware components, the implementation of the software logic and its {interactions with the} dynamic environment. Achieving this {multi-modal} integration is crucial for performing complex tasks requiring learning based on sensorimotor skills \cite{wolpert2011principles}.

%As the AI systems begin to replace many critical positions of human occupancy in making mission-critical and life-changing decisions, ensuring that they are safe, secure, and robust against adversarial attacks, system failures, and unintended intrusions gradually becomes a major focus of today's research. This includes developing techniques for verifying and validating AI behavior and mitigating potential risks associated with their deployment. Maybe the missing elements could be as far-stretched as developing self-awareness and characterization of self-assessments. The conceptualization of consciousness and self-awareness in AI systems remains one of the most widely studied and elusive goals \cite{lewis2011survey, chella2020developing}. Investigating the nature of consciousness and exploring whether it is possible to create artificial entities that possess subjective experiences such as feeling secure or safe, similar to humans, is still largely unknown. Although there have been attempts at mathematical modeling \cite{kleiner2020mathematical, namazi2012mathematical}, these have not yet sufficiently explained the complete conscious experience that humans present.

\vspace{-3mm}
\subsection{Transferability of human-specific attributes to machines}

To transfer a uniquely human attribute—-such as ethics, aesthetics, consciousness, morality or trust—it is essential to define these attributes using clear and specific language. Without a clear systematic description, it would be {challenging} to have a common consensus on their meanings.  However, creating a {common ground} entails improving the specificity of their mathematical definitions \cite{woodford2020modeling}.  The difficulty of creating mathematical definitions is partly due to {the} noisy nature of their evolutionary development. In fact, the imprecision in the development of such human attributes can be shown to be beneficial in accurate and robust decision--making processes.

%In fact, having a specific definition for biological and metabolic events could sometimes be pretty hard to frame, which are analyzed historically using a branch of mathematics known as information theory \cite{stone2018principles}. 

\subsubsection{The role of imprecise nature of perception in learning}

{Humans have evolved to rely on approximate reasoning to navigate uncertainty, enabling flexible responses to imprecise or ambiguous information. In interpreting social cues, making ethical judgments, or appreciating humor, they often operate through subjective interpretation rather than absolute precision. Such traits—ethics, aesthetics, humor, and others—are integral to human cognition and shape how people perceive and interact with the world \cite{findling2021imprecise, franke2018vagueness}.} There is a compelling rationale for incorporating the modeling and processing of imprecise information into learning and expert systems. By doing so, we can better emulate the complex, layered, and often ambiguous ways in which human reasoning operates. {For example, embedding the ability to handle imprecise information in AI systems, such as virtual assistants, can enhance their responses to human emotions, support ethical decision-making, and generate humor more naturally and relatably.}

%Human beings have evolved to rely on approximate reasoning as a fundamental strategy for navigating the inherent uncertainties of the world such as found in volatile environments \cite{findling2021imprecise}. This capability allows them to process and respond to imprecise, incomplete, or ambiguous information with remarkable flexibility and adaptability. For instance, when interpreting social cues, making ethical judgments, or appreciating humor, humans often operate within a realm of subjective interpretation rather than absolute precision. These traits—ethics, aesthetics, humor, and beyond—are not merely ancillary to human cognition but are deeply intertwined with the very nature of how humans perceive and interact with the world \cite{findling2021imprecise, franke2018vagueness}.

Moreover, embracing the nuances of imprecision in AI models aligns with a broader movement in cognitive science that recognizes the value of ``fuzzy" logic and probabilistic reasoning as closer approximations to human thought processes \cite{qu2019probabilistic}. {This approach not only enhances AI robustness in real-world complexities but also enables new directions such as multi-modality model development and multitasking. In summary, modeling imprecision seeks not merely to replicate human reasoning but to enrich AI’s capacity to engage with the subtle dimensions of 3D world models and human experience.} %This approach not only makes AI systems more robust in dealing with real-world complexities but also opens new avenues for creating machines that can truly resonate with the multifaceted nature of human experience such as multi-modality model development and multitasking. In summary, by modeling imprecision, we do not just aim to replicate human reasoning but to enrich AI's capacity to engage with the subtle and sophisticated dimensions of 3D world model and human experience roaming within. %pearl2014probabilistic

\subsubsection{The effect of neuroscience studies for real-world stimuli}

%\textcolor{red}{Most experiments take place in highly controlled lab environments. However, incorporating real-world stimuli and designing experiments with multisensory signals and various confounds can create more realistic conditions. This approach can help us better understand the types of inputs the human brain processes daily.}

{Traditional neuroscience experiments are typically conducted in highly controlled laboratories. While such settings are invaluable for testing isolated hypotheses and uncovering fundamental neural mechanisms, they often fail to capture the complexity of real-world experiences. Incorporating real-world stimuli—through multisensory signals and diverse confounds—can therefore enhance the ecological validity of neuroscience studies \cite{shamay2019real}.} %francken2018neuroscience

%In traditional neuroscience research, most experiments are conducted within the confines of highly controlled laboratories. While these controlled settings are invaluable for formulating isolated hypotheses and understanding fundamental neural mechanisms, they often fall short of capturing the complexity and variability of real-world experiences. To bridge this gap, it is recommended that incorporating real-world stimuli into experimental designs—integrating multisensory signals and introducing various confounds—can significantly enhance the ecological validity of neuroscience studies \cite{shamay2019real,francken2018neuroscience}.

By designing experiments that more closely mirror the diverse and dynamic conditions encountered in everyday life, researchers can gain a deeper understanding of how the brain processes the rich tapestry of sensory information received on a daily basis. {For example, instead of presenting isolated visual or auditory cues, experiments can immerse participants in VR/AR scenarios that combine synchronized sights, sounds, and tactile feedback—more closely approximating real-world interactions and yielding richer data for Brain-Computer Interfaces (BCIs) \cite{wen2020current}. This approach provides insights into how different sensory modalities interact in the brain and reveals how contextual factors, such as distractions or emotional states, influence perception and decision-making.}  %This approach not only provides insights into how different sensory modalities interact in the brain but also reveals how contextual factors, such as distractions or emotional states, influence perception and decision-making. % BCI:  krugliak2022towards

Furthermore, studying the brain's response to realistic stimuli can uncover how neural processes adapt to unpredictable environments. {Such research directly confronts higher-order cognition—attention, memory, problem-solving—constantly tested by the richness of real-world demands. Escaping the constraints of impoverished lab tasks allows neuroscience to uncover how the brain so admirably winnows, combines, and deciphers the information we are confronted with daily.} %This line of research is particularly relevant for understanding higher-order cognitive functions such as attention, memory, and problem-solving, which are constantly challenged by the multifaceted nature of real-world tasks. By moving beyond the artificial simplicity of lab-based experiments, neuroscience research can begin to unravel the brain's remarkable ability to filter, integrate, and make sense of the overwhelming amount of information encountered daily in natural settings. 
{Ultimately, this shift towards realistic experimental paradigms could transform our understanding of brain function, revealing how neural circuits support adaptive behavior under continuous multi-modal sensory input. This may enable more effective interventions to enhance cognitive function in daily life, from educational tools to BCIs \cite{9919502, wen2020current}.}

%Ultimately, this natural shift towards more realistic experimental paradigms holds the potential to transform our basic understanding of brain function, offering new insights into how neural circuits support adaptive behavior in the real world where the brain is constantly exposed to multi-modal sensory input. Such advances could pave the way for more effective interventions and technologies designed to support cognitive function in everyday life, from improving educational tools to developing brain-machine interfaces  that operate seamlessly within the complexities of the real world \cite{9919502, wen2020current}.

\begin{figure}[t!]
    \centering
    \includegraphics[width=0.95\columnwidth]%{benchmark3_notitle.pdf}
    {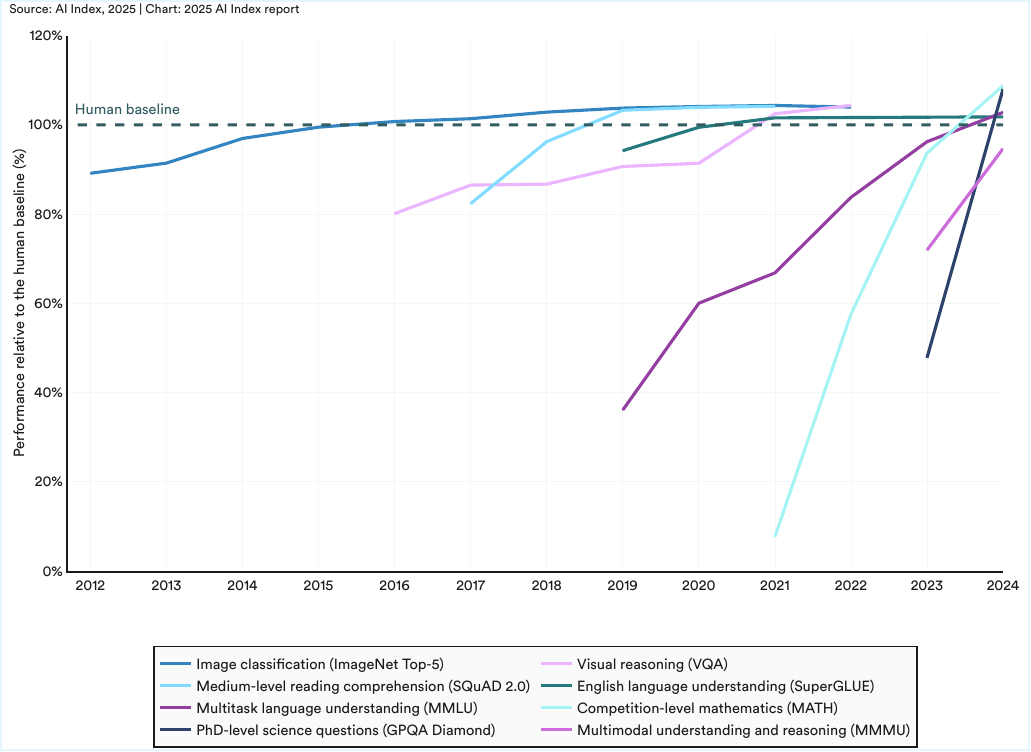}
    \caption{\textcolor{gray}{Select AI Index technical perfrmance benchmarks v.s. human performance. Source: AI Index, 2025 (AI Index Report). Available from: \url{https://aiindex.stanford.edu/report/}}}
    \label{fig:Benchmark}
\end{figure}

\subsection{Alignment of SoTA approaches with HIns/HAss AI}

Today's AI models exceed humans in a  {number} of intellectual tasks, as outlined in Fig. \ref{fig:Benchmark}. Yet, these models are useful in closed and regulated application domains, successfully addressing  the specific task requirements \cite{samaranayake2025transformative}. There are various angles of criticism, including algorithmic bias, ethical concerns, safety issues, over-reliance, impact on creativity development, among others \cite{hagendorff202015}. %Rather than diving into the details of each critical point, we will rather explore the alignment of current models in the context of hardware, architecture built on {or around} this hardware, and the learning schema that generates the software in the following subsections. 
{Instead of detailing each critical point, we examine the alignment of current models with hardware, the architectures built around it, and the learning schemas that generate the software in the following subsections.} We shall start considering the effect of the today's hardware architectures on AI system development.

\subsubsection{The influence of the architecture of today's computational models} \label{VNeffectonAI}

Let us begin {by considering}
 Marr's three levels which states that any complex information processing machine must be interpreted through the lens of three levels of organization \cite{marr2010vision}: 
\begin{itemize}
    \item \textbf{\textit{Computational theory}:} Goal and appropriateness of the computational logic or strategy. 
    \item \textbf{\textit{Representation and algorithm}:} Representation for the {I/O} and how the algorithm transform one to the other. 
    \item \textbf{\textit{Hardware and implementation}:} The physical implementation of the algorithm.
\end{itemize}

Marr's approach argues that complex systems {form} a hierarchy where each level consists of composites of the entities at a lower level. These systems can be interpreted at various levels, {e.g.} biochemical, cellular, psychological in organisms or logical gates, machine language, {functionals} in computers. Higher-level explanations allow for generalizations across different underlying physical structures. This means we can predict behaviors without detailed knowledge of the underlying components, like predicting genetic traits without knowing DNA specifics or predicting a computer's output without understanding the properties of semiconductors. The power of higher-level explanations lies in their ability to offer reasonable predictions with less detailed information about the system's behavior. {They attain generality by emphasizing subsystem input–output functions rather than the significance of their internal mechanisms.} %They achieve generality by focusing on the input/output functions of subsystems rather than the functional significance of their inner workings.

%The success of today's AI heavily depends on the data representations and algorithm developments just like we see it in the development of neural networks and the algorithms that enabled such computational models become realistic such as Bayes updates, gradient descent, dynamic and probabilistic methods and theories. 
{The success of modern AI largely depends on data representations and algorithmic advances, as seen in neural networks enabled by methods such as Bayesian updates, gradient descent, and probabilistic approaches.} The main medium is silicon-based transistor logic employing Von Neumann architectures, characterized by large, inactive memory and minimal state within the central processing unit. Due to the significant bottleneck posed by solo processors and slow memory for real-life workloads, parallel architectures have garnered traction, enabling independent logic to be executed concurrently. %Consequently, numerous AI model development endeavors appear to be gravitating towards maximizing the efficiency of parallel architectures for training. A notable instance of this trend is the recent advancement in transformer models, which exhibit a high degree of parallelizability compared to convolutional neural networks \cite{vaswani2017attention, shoeybi2019megatron}. 
{Consequently, many AI development efforts focus on enhancing the efficiency of parallel architectures for training. A prominent example is the transformer models, which are far more parallelizable than CNNs \cite{vaswani2017attention,shoeybi2019megatron}.} Obviously, such a bias leads to a broad range of changes in the way the algorithms are designed, ultimately changing the implementation details of intelligence on silicon. % 

Decoupling the implementation from the set of required computations for visual perception should enable the development of human-like AI on silicon without any fundamental limitations. However, this assumption hinges on a complete understanding of the computations needed for visual perception. {Once the computational steps underlying perception are fully understood, they could be implemented to achieve human-like interpretations. Yet the biological computation framework remains unresolved, and we often depend on neural models to study complex systems. Consequently, interpretations of biological learning are closely tied to hardware design, including architecture, power use, and data handling.}

%Once the functionally realizable computational steps governing perception are fully elucidated, we should be able to implement these steps with sufficient instructions to achieve high-level human-like interpretations. Nevertheless, the biological computation framework remains unresolved, and we frequently rely on computational models to understand complex biological systems. Hence, our interpretations of biological hardware and how learning transpires become unequivocally dependent on the way hardware is manufactured, including the architecture, power use and data handling. 

%One key distinction between biological and machine intelligence is the strength of the coupling between the software implementing the aspects of intelligence and the available hardware this software is designed and built on. Abstraction between the layers in any complex system allows modularity between system components, making design manageable but complicating debugging and failure tracing. Additionally, as the data volume increases and the models get chunkier,  the implementation of non-biological (machine) intelligence seems to require significant resources, exponentially scaling the required power to support their learning. However, machine intelligence, with its parameter values, network links, and overall model descriptions, is highly efficient for knowledge storage and transfer across different digital hardware, reflecting the decoupled design aspect. 

{One of the key distinctions between biological and machine intelligence lies in the hardware-software pairing. Abstraction across layers allows modular construction but makes debugging and tracing failures more difficult. With more data and models, machine intelligence needs exponentially more resources for learning and yet needs less in terms of storing and shipping knowledge across digital hardware, reflecting its decoupled nature. In otherwords, the resource-intensive learning process can be executed once and then transferred or specialized. By contrast, human intelligence is tightly coupled with biological hardware, which provides physiological advantages but requires each system to be trained from scratch with innate knowledge architecturally embedded and less transferable.}

%In otherwords, the learning process, despite being heavy and overwhelming, could be executed only once for all and then transferred/distributed for further learning through specializations, which also highlights the significance of foundation models. In contrast, human intelligence is tightly coupled with biological hardware, offering physiological advantages besides its efficient analog knowledge representation and processing. This tight coupling makes the overall learning process quite challenging, as each hardware needs to be trained from scratch using innate knowledge integrated into the architecture. 

\subsubsection{Model Architecture and Learning}

\paragraph{Model Architecture} In the evolving landscape of modern ML models in CV, CNNs and the Vision Transformer (ViT) stand out as two of the most impactful approaches, each offering distinct architectural and representational features \cite{raghu2021vision}. While CNNs have long dominated tasks like object detection and image classification, recent advancements in (multi-head) attention-based Transformer architectures, particularly ViT \cite{dosovitskiy2020image}, challenge the conventional reliance on convolutions in CV. Despite the recent work \cite{cordonnier2019relationship} on the expressive equivalence of naive convolutions and the implementation of attention mechanisms, the advent of transformer's internal components points at alternative feature extraction techniques which may or may not align with the brain's intricate processing of objects. % (which is akin to {the mechanics} of convolution operation). 

{CNNs have long been the standard for CV tasks, with much effort devoted to understanding their representations on datasets like ImageNet \cite{deng2009imagenet}. Diagnostic datasets such as Stylized ImageNet \cite{geirhos2018imagenet} and CIFAR10-c/100-c \cite{hendrycks2019benchmarking} have further revealed model biases—for instance, large pre-trained CNNs tend to favor texture over shape, unlike human categorization \cite{geirhos2018imagenet}. Like human vision, CNNs excel at capturing local patterns and show translation invariance, making them robust to changes in object position and orientation. Their layered architecture is also somewhat interpretable: early layers capture simple features (edges, textures), while deeper ones represent complex and discernible patterns \cite{lindsay2021convolutional}. Although weight sharing improves memory efficiency, optimizing over the entire image space remains restrictive, and fixed input size is yet another limitation.} The static nature of CNN's layered structure and optimization (learning) are nowhere near how the brain operates, and is {developmentally} structured and trained. %trapp2015human

%Convolutional Neural Networks (CNNs) have been the de facto standard for a range of computer vision tasks, and significant efforts have gone into understanding their learned representations on popular datasets like ImageNet \cite{deng2009imagenet}. The emergence of diagnostic datasets, such as Stylized ImageNet \cite{geirhos2018imagenet} or CIFAR10-c and CIFAR100-c \cite{hendrycks2019benchmarking}, has deepened our comprehension of model biases. For example, it is shown that most of the pre-trained large scale CNN models are biased towards texture rather than shape unlike human categorization processes \cite{geirhos2018imagenet,trapp2015human}. As is the case with the human vision, CNN's convolutions are adept at capturing local patterns in images and exhibit translation invariance, making them robust to variations in position and orientation of objects within the image. In addition, CNN's layered architecture can be considered more interpretable with respect to functional organization of human visual system, as features learned in earlier layers often correspond to simple patterns like edges and textures, while higher layers capture more complex features \cite{lindsay2021convolutional}. Despite parameter (weight) sharing is more memory efficient, optimizing parameters for the entire image space can be seen as quite restrictive, not to mention the fixed input size. The static nature of the layered structure as well as the optimization (learning) are nowhere near how the brain operates, and are {developmentally} structured and trained. 

{Recently, the landscape has shifted with the emergence of Transformer-based variants such as the ViT \cite{dosovitskiy2020image,liu2021swin}, demonstrating notable flexibility and effectiveness across domains including time-series analysis and natural language processing. These architectures now serve as the foundation of large-scale models \cite{bommasani2021opportunities}, supported by distributed computational infrastructures. ViTs capture global dependencies in visual data by representing images as sequences of patches, thereby enabling more effective modeling of long-range interactions. This patch-based representation is inherently scalable, as high-resolution images can be processed without recursive compute cycles, while workload independence makes the approach fit for parallel execution.}

%However, the landscape has evolved with the introduction of various tweaks for Transformer architectures like  Vision Transformer (ViT) \cite{dosovitskiy2020image,liu2021swin}, known for their flexibility and success in time-series data analysis and natural language processing tasks. They now form the fundamentals of foundation models \cite{bommasani2021opportunities} with large scale distributed hardware support. ViTs capture global dependencies in images by treating them as sequences of patches, allowing them to potentially model long-range interactions more effectively. Such patch-by-patch processing could be quite scalable as high-resolution images can easily be learned without multiple runs of tedious processing and plus workload independence lends this processing to be easily executed on parallel architectures. 

%Recent models have key components that make them unique in the learning trajectory such as Convolution vs. Attention: While CNNs demonstrate power and scalability, their reliance on local connectivity raises concerns about global context loss. Transformer models, rooted in attention mechanisms, provide an alternative that challenges this paradigm. 
%The introduction of Transformer architectures to computer vision adds a layer of complexity and adaptability, assisting vision models to infer long range dependencies more effectively. 
{Transformers in CV enhance adaptability and enable more effective modeling of long-range dependencies.} Superiority of Transformers is largely  credited to their self-attention-like architecture which might not necessarily be biologically plausible. {Despite limited biological plausibility, such architectures dominate benchmarks due to scalability, parallelism on current hardware, and effective long-range modeling.} There have been many attempts to improve the effective receptive field size of these networks to compensate for the long range dependencies \cite{ding2022scaling}. In fact, in \cite{wang2022can}, authors highlight three highly effective architecture designs for boosting robustness: (1) patchifying input images, (2) leverage of larger kernels and (3) reducing activation/normalization layers. Although attention-based activations are {general} and can implement convolution, their presence seems to be {optional} to ensure robustness in various tasks of categorization and recognition. %\cite{li2021can}
These findings  {collectively} indicate that there are likely several architectures for specific biologically plausible functionalities. Indeed, a few published works already argue {that} biological plausibility  i.e., neurons can perform attention-like computations through short-term, Hebbian synaptic potentiation \cite{ellwood2024short}. %When an axon's activity matches a neuron's activity, the synapse is briefly potentiated, enabling the axon to control the downstream neuron. 

{Optimized inference techniques, such as DeepSeek’s hybrid bottom-up query compression and top-down adaptive caching, reflect human perceptual hierarchies \cite{liu2024deepseek}. These methods merge raw sensory-like data processing (bottom-up) with knowledge-driven filtering (top-down), similar to how the brain integrates sensory input with context. In addition, the Mixture of Experts and its gating network act as a top-down controller, dynamically selecting the right subset of experts—a computational analog of attention in biological systems.}

%Optimized inference techniques, such as \textbf{DeepSeek}’s hybrid bottom-up query compression and top-down adaptive caching, genuinely reflect human perceptual hierarchies \cite{liu2024deepseek}. These methods combine raw sensory-like data processing (bottom-up) with prior knowledge-driven filtering (top-down), paralleling how the brain integrates sensory input with contextual expectations. In addition, previously elaborated Mixture of Experts and the gating network acts as a top-down controller, dynamically selecting the right subset of experts—a computational analog of attentional mechanisms in biological systems.

\paragraph{Model Learning} {Evolution as a computational method for biological optimization uses evolutionary algorithms to tackle complex, often constrained problems. These algorithms mimic processes like mutation, crossover, and selection to gradually improve solutions, applying biological principles to find optimal or near-optimal configurations in domains such as genetics, neural networks, and ecology. By leveraging the efficiency and adaptability of evolution, this approach addresses intricate challenges in biochemical mechanisms and computational biology.}

%Evolution as a computational approach to biological optimization refers to the application of evolutionary algorithms to solve complex and often constrained optimization problems. These algorithms mimic the mechanisms of mutation, crossover, and selection to iteratively improve solutions, leveraging biological principles to find optimal or near-optimal configurations in various domains, including genetics, neural networks, and ecological modeling. This approach capitalizes on the inherent efficiency and adaptability of evolutionary processes to address intricate challenges in biochemical mechanisms and computational biology.

Although BP is not necessarily {the right direction} to attain biological plausibility \cite{mazzoni1991more}, it is a good idea to leverage the main properties of ecosystems in building computational tools to solve complex problems \cite{parpinelli2012biological}. The possibility of inserting ecological elements in the optimization process is {important} in the development of new biologically plausible hybrid systems. 

\begin{figure}[t!]
    \centering
\includegraphics[width=0.55\columnwidth]{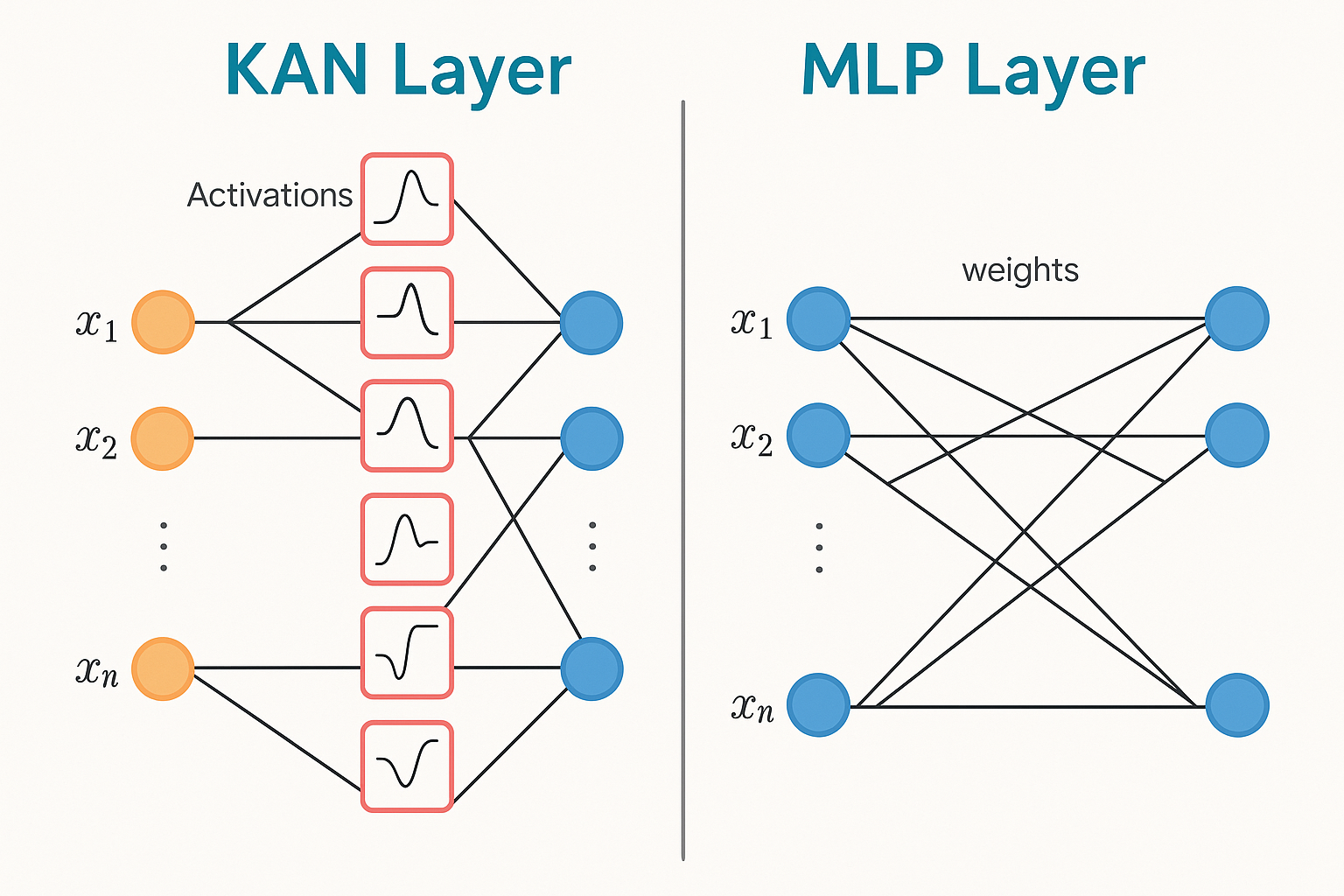}
    \caption{\textcolor{gray}{Comparison between MLP v.s.  KAN layer. While the edges of MLP just hold one weight, the edges of KAN hold more parameters because of splines.}}
    \label{fig:MLPvsKAN}
\end{figure}

\subsubsection{Comments on building blocks (MLP, KAN and beyond)}

%\textcolor{red}{This section is dedicated to discuss the origins of Multi-Layer Perceptron model and universal approximation theorem. Most recently Arnold-Kolomogrov theorem is invokved to design Arnold-Kolomogrov Networks. Both approaches are computational hardware interpretations that emulate biological information transfer and processing that takes place down at the edges: synapses and axons/dendrites.}

The Multi-Layer Perceptron (MLP) \cite{pal1992multilayer} is a widely adopted model to simulate biological information processing through continuous approximation mechanisms. In other words, it is a crude representation of high-level functionalities of synapses and axons/dendrites in the human brain, making it easier to apply calculus rules and to implement learning on digital hardware with the available instruction sets. The generalized use cases of MLP consists of multiple layers of interconnected nodes, or so called neurons, where each neuron performs a weighted linear sum of its inputs followed by a non-linear activation function as shown in Fig. \ref{fig:MLPvsKAN}.  Universal Approximation Theorem (UAT) \cite{kidger2020universal} enables MLPs to approximate any arbitrary non-linear continuous function by learning and modeling complex patterns and relationships.

%Arnold-Kolmogorov (AK) theorem suggests that any multivariate function can be decomposed into a sum of compositions of univariate functions, offering a neat theory for designing Kolmogorov-Arnold Networks (KANs) \cite{liu2024kan}. KAN networks leverage the AK theorem to simplify the construction of neural networks capable of approximating complex functions with fewer parameters and potentially more efficient architectures in which weightening is replaced with individual spline-based activation functions (see Fig. \ref{fig:MLPvsKAN}).

{The Arnold–Kolmogorov (AK) theorem shows that any multivariate function can be expressed as a sum of univariate functions, providing the foundation for Kolmogorov–Arnold Networks (KANs) \cite{liu2024kan}. KANs apply this principle to approximate complex functions with fewer parameters, replacing traditional weighting with spline-based activations for potentially more efficient architectures (see Fig.~\ref{fig:MLPvsKAN}).}

{UAT and AK theorems offer alternative approaches to multivariate function approximation and have inspired hardware designs aimed at emulating biological information processing. Correspondingly, MLPs and KANs represent related but distinct architectures: MLPs emphasize linearity, flexibility, and generalization with large datasets, whereas KANs exploit edge non-linearity to optimize representation and efficiency. Although KANs may reduce parameter counts, two observations follow: (1) MLPs with learnable activation functions expressed as spline sums can approximate KAN activations; (2) this equivalence implies that challenges in MLPs, such as the curse of dimensionality, would naturally extend to KANs.}

%UAT and AK theorems are two alternatives for multivariate complex function approximations and have found distinct implementation styles, inspiring computational hardware designs that might emulate biological information processing better. Concordantly, MLPs and KANs represent two different but similar approaches to neural network architecture design. While MLPs focus on signal linearity, flexibility, and generalization capabilities with voluminous data, {KANs} aim at non-linearity at the edges, optimizing the representation and computational efficiency. Despite the potential of reducing the number of parameters, theoretically, we can make a few observations: (1) MLPs with learnable activation functions that could be formulated in terms of the sum of splines have the potential to approximate activations of KANs. (2) This theoretical equivalence leads to the conclusion that the main issues with MLP-based network design such as the curse of dimensionality are still applicable to the fundamentals of KANs. 

%On the other hand, theoretical equivalence does not necessarily {imply that} the practical implementation of these networks is the same in terms of power usage, parameter optimizations, training time, implementation complexity, etc. 
{Theoretical equivalence does not necessarily translate into identical practical implementations with respect to power consumption, parameter optimization, training and implementation complexities. For instance, hardware implementations optimized for spline activations allow hardware to handle non-linearity while software executes higher-level linear sums, enabling digital-friendly complex operations. These instruction-level couplings may further optimize power consumption, transistor counts per bit, and response latency.} 

%For instance, let us think about hardware implementations optimized to perform spline activations, and hence while the hardware deals with the non-linearity, the software can run higher-level simpler operations such as linear sums, enabling complex operations to run closer to digital hardware through creating close couplings. These instruction-level couplings could be a source of further optimizations in terms of power consumption, transistor counts per processed bits or {response} latency etc. 

\subsubsection{Texture/Shape Bias and Fine-Tuning}

An exploration into shape bias reveals that ViT  tends to align more closely with human-like error consistency when fine-tuned \cite{dosovitskiy2020image}. In contrast, ResNet exhibits a {less} human-like error distribution. Data augmentation reshapes model representations (via stylized datasets), impacting their similarity to human behavior. While Transformers, e.g., ViT, showcase superior accuracy, and exhibit a higher shape bias, contrary to most CNNs usually being texture--biased \cite{wang2022can,raghu2021vision}. %naseer2021intriguing 
%The fine-tuning of models with augmented data further influences their error consistency, with ViT trending towards a human-like pattern. 
{ViT trends behave like a human, which} {could be due to} diminished influence of inductive bias. However, a drawback arises with ViTs \cite{han2022survey} as they necessitate a significant volume of data for training to achieve {competitive accuracy}, contrasting sharply with the lower data requirements in humans.

\section{Final Thoughts and Discussions}
\label{section6}

{To elaborate on the strategic role of humans in AI development, we primarily need to analyze the sources of current AI hype. From our observation, there are at least two major domains where AI competes with and sometimes surpasses human performance: (1) Controlled tasks/applications, such as DeepMind’s AlphaGo \cite{silver2017mastering}, powered by reinforcement learning, operates in closed-ended domains, relying on surface correlations and rule-based equations that can be reasonably defined. This resembles earlier ``first-generation" AI \cite{zhang2023toward}, where symbolic objects were used to model realism and solve problems within defined boundaries. Unlike classical reinforcement learning, RLHF makes humans part of the reward loop, enabling FMs to internalize complex, subjective human preferences that cannot be captured by fixed reward functions and widening their applicability beyond rule-based domains. (2) Lower-dimensional realisms or symbolic formal systems, found in human constructs like language, drawings, and other expressive forms of thought and emotion. These cognitive models are projections of external realism—a highly complex, multi-dimensional process that may be impossible to fully model. Literature often distinguishes between closed-ended domains (e.g., rule-based games, object and speech recognition) and open-ended domains, where intelligence develops into common sense.}

{As we move to lower-dimensional projections of 3D realism, objects within it (e.g., words, numbers) show more correlative structure or discernible patterns, making them easier to conceptualize as “correlations.” This risks misunderstandings that can appear as spurious correlations \cite{vigen2015spurious}. In simplifying complex phenomena and their brain representations, we reduce dimensionality to increase surface correlations. This, however, leads to the removal of causality, common sense, counterfactual reasoning, and comprehensive perception—e.g., the higher-dimensional understanding of realism, interactions and internal models of external phenomena.}

%One observation is that as we go to lower dimensional projections of our 3D realism, objects living within that realism (e.g. words, numbers, etc.) seem to show more correlative structure or discernible/sensible patterns which naturally become easier to conceptualize and reduce down to the concept of "correlations". This has the danger of leading to misunderstandings which might manifest itself in the form of spurious correlations \cite{vigen2015spurious}.  In other words, to simplify rather complex phenomena and its representations in the brain, we concisely interfere and make an attempt to reduce the dimensionality to increase the ``surface" correlations.   This ultimately lends itself to the absence of common sense, counterfactual reasoning and comprehensive perception --like the 3D or higher dimensional understanding of realism and interactions within, intertwined with causal reasoning and development of internal models around the external phenomena.

%Despite this observation, outstanding performance of the computational frameworks led many research groups to heavily rely on the idea that incorporating human expertise into machine learning systems could be undesirable for introducing bias and, to some extent, dishonesty \cite{zimmerman2023human}. 

{Despite this, the strong performance of computational frameworks led many researchers to view incorporating human expertise into ML as undesirable, potentially introducing bias or  dishonesty \cite{zimmerman2023human}. In fact, they proposed that a sizable portion of the ML community would be greatly aided if common sense could be inferred from large-scale pleathora of research with minimal prior information from a human model.} This trend might be the reason for the emergence of self-supervised, self-motivated or self-improving ML  \cite{omohundro2007nature}. This new trend represents the sheer desire of many AI leaders to search for alternative intelligences that could have better shortcuts compared to the evolutionary trajectory of human intelligence development for convergence to an equilibrium: Reaching a level of \textit{super-human intelligence}. {Conversely, RLHF illustrates how human involvement can improve alignment and safety without stifling scalability, highlighting that reducing human input might not be the only viable trajectory.} 

{We hypothesize that bio-inspired AI has greater potential to be ethical, fair, and less judgmental. Models rooted in biological systems, whose functional anatomy is well understood, may naturally embody fairness and reduced bias through their emergent behavior. This assumes that a system’s architecture and functions shape its useful emergent traits. Drawing from biology, such models might better align with ethical principles. A key reason is their relative comprehensibility: interpretability helps integrate fairness, ethics, and human values.} %Since bio-inspired models build on well-studied biological systems, their functional anatomy is better understood, making their behavior easier to interpret and ethical considerations easier to grasp and embed.}

%We could also formulate a hypothesis that bio-inspired AI has a better potential to be more ethical, fair and less judgmental. This line of thinking suggests that AI models derived from biological systems might be better equipped to embody ethical behavior, fairness, and reduced bias compared to other AI forms as part of their emergent behavior. When making such a claim, we assume that the architectural and functional properties of a complex system have a direct impact on all the useful emergent behaviors it generates. Drawing inspiration from biological systems, these models may align more closely with ethical principles, exhibiting less judgmental behavior. Several reasons support this stance. The foremost is the comprehensibility and interpretability of such models: enhanced interpretability aids in integrating fairness, ethics, and human values. Biologically plausible or bio-inspired models, grounded in well-explored biological systems, benefit from extensive studies that unravel their functional anatomy. This inherent comprehensibility simplifies the interpretation of their roles in various contexts, thereby facilitating the integration of ethical considerations.

{Human values arise from emergent behaviors shaped by the brain’s unique anatomy and neural structure, so it’s not surprising to expect similar traits in bio-inspired systems.} This idea holds the view that the behavioral similarity could be coupled with the endogenous structural similarities. {While other forms of intelligence may also show such traits, the chances are lower. Put simply, similar structures may yield similar functions under certain conditions—though research is needed to define those conditions and clarify how architecture couples with algorithms and applications.}

%Additionally, human values stem from emergent behaviors within the human brain's peculiar anatomy and the unique structural characteristics composing the neural system. Therefore, it shouldn't be surprising to anticipate similar behaviors emerging from bio-inspired systems. This idea holds the view that the behavioural similarity could be coupled with the endogenous structural similarities. While it's plausible for other forms of intelligence to exhibit these human traits, the likelihood of such emergence might be comparatively lower. In other words, it seems reasonable to hypothize that similar structural properties might lead to similar functional and operational characteristics under certain conditions. The genuine research work is needed to precisely define such conditions and demonstrate the inherent coupling between the hardware (the architecture) and the software (algorithm and/or learnable applications). 

Bio-inspired systems inherently possess a greater level of inductive bias, {requiring} fewer data samples for learning. While this might appear to limit the overall learning process, human values like fairness or ethical behavior can be guided by a limited number of examples throughout a lifetime, suggesting that the innateness could be driven largely by the structural properties. {In essence, such values can be embedded by adjusting inductive bias and making structural changes to the architecture\footnote{DeepSeek V2/V3 \cite{liu2024deepseek} showed how bio-inspired principles—modularity (MoE), memory efficiency (MLA), and hierarchical processing—reduce memory overhead and improve scalability. These brain-like parallels enable robust, energy-efficient AI without directly replicating biology.}. In bio-inspired models, these modifications are generally easier than in systems built on entirely different forms of intelligence.}

\section{Conclusion}
\label{concSection}

%\section*{Acknowledgment}

%Reflecting on the strategic placement of humans in AI development, it becomes critical to examine the origins of the current AI hype. A closer look, as presented in this paper, reveals two distinct domains where algorithms has not only rivaled but, in some cases, surpassed human performance. The first domain is characterized by controlled tasks, such as those mastered by DeepMind’s AlphaGo, where reinforcement learning has thrived within closed-ended systems, exploiting well-defined surface correlations and rule-based frameworks. 
{Reflecting on the strategic role of humans in AI development, it is critical to examine the origins of current AI hype. This paper highlights two domains where algorithms have rivaled—and sometimes surpassed—human performance. The first involves controlled tasks, such as DeepMind’s AlphaGo, where reinforcement learning excelled in closed systems by exploiting surface correlations and rule-based frameworks.} This echoes the earlier, symbolic AI approaches that tackled intelligence within similarly constrained realisms. {The second domain concerns lower-dimensional realities, where human constructs like natural language or symbolic systems—seen as projections of the 3D world—simplify complex phenomena into patterns. Yet in higher-dimensional, open-ended domains like common-sense reasoning, the difficulty of modeling becomes clear, exposing the limits of current AI, especially FMs.}  %The second domain is related to lower-dimensional realities, where human constructs like natural language or symbolic formal systems—often perceived as projections of the external 3D world—simplify complex phenomena into discernible patterns. However, as we transition to higher-dimensional, open-ended domains like common-sense reasoning, the complexity of modeling these realities becomes apparent, highlighting the limitations of current AI methodologies, most starkly the foundation models.

{Although many recent advances in computational frameworks take inspiration from human cognition and perception, there is a growing push to reduce human involvement in AI development, largely due to concerns about bias. This has led to self-supervised learning, aimed at achieving forms of intelligence that could even surpass humans. Yet, bio-inspired AI presents an appealing alternative. By leveraging the structures and functions of biological systems, it promises models that are more interpretable, ethical, fair, and inherently aligned with human values. Thanks to their strong inductive biases—closely matching how the real world is structured—these systems may require fewer data samples to learn ethical behaviors.}

%Despite the remarkable advances in computational frameworks mainly inspired by human cognition/perception, there has been a recent growing tendency to minimize human involvement in AI development, driven by growing concerns about introduced bias. This trend has led to the rise of self-supervised learning approaches, to achieve super-human intelligence that could potentially outpace the evolutionary trajectory of human cognition. Yet, amid this shift, bio-inspired AI emerges as a compelling alternative. By drawing on the structural and functional insights from biological systems, bio-inspired AI offers a pathway to more informed, ethical, fair, and interpretable models. These systems, grounded in well-explored biological frameworks, are shown to align with human values inherently and may require fewer data samples to learn ethical behaviors due to their significant and more aligning (with the 3D world model) inductive biases.

{The hypothesis that similar structures yield similar functions underscores the promise of bio-inspired AI, where the link between architecture and functionality aligns more closely with human cognition. This ultimately makes it easier to embed ethical and moral considerations. As debates continue, bio-inspired approaches stand out as a promising avenue to balance computational efficiency with the human-specific attribute imperatives that should shape the path towards AGI.}

%The hypothesis that similar structural properties could lead to analogous functional outcomes underscores the potential of bio-inspired AI. The coupling between architectural design and algorithmic functionality in these models suggests a closer alignment with human cognitive processes, offering a more natural integration of ethical, aesthetic, and moral considerations. As the debate continues over the future direction of AI technology, bio-inspired approaches present a promising avenue for bridging the gap between computational efficiency and the human-specific attribute imperatives that must guide the evolution of the efforts for AGI. 

%\section{Acknowledgement}

%The ideas presented in this document stem from a reassessment of my interpretations of the human role in the vast domain of intelligence, informed heavily by local talks at MIT (such as those at the Simons Center for the Social Brain, the Center for Brains, Minds \& Machines, etc.), interactions with the faculty as well as researchers at vision conferences (VSS, etc.), technical literature readings, and collaborative discussions initiated with experts at the Sinha Lab. I extend my deepest gratitude and appreciation to all those involved in these enlightening exchanges.

%\section*{References}

\bibliographystyle{ieeetr} %alpha, apalike, ieeetr
\bibliography{bibliography.bib}

%\begin{IEEEbiographynophoto}{Suayb S. Arslan}{\space}(M'07--SM'22) received the M.Sc. and Ph.D. degrees in Electrical and Computer Engineering from the University of California, San Diego, CA, USA, in 2009 and 2012, respectively. Dr. Arslan has worked as an R\&D engineer for MERL in 2009, Cambridge, MA and senior researcher for Quantum Corporation, Irvine, CA between 2012 and 2016 and has been part of various projects involving object tracking and persistent data storage. He is currently affiliated with the Department of Brain and Cognitive Sciences at Massachusetts Institute of Technology, Boston, MA, USA and serves as a professor in the Department of Computer Engineering at Bogazici University, Istanbul, Türkiye. He received numerous recognitions including Fulbright grant and outstanding research awards from different scientific institutions.  His research interests include information theory, neuroscience, digital communication, networking and storage, Cloud and Quantum computing, reliability/system theory and image/video processing. He is an associate editor for Elsevier IoT Journal. 
%\end{IEEEbiographynophoto}

\end{document}